\documentclass[10pt,twocolumn,letterpaper]{article}

\usepackage[pagenumbers]{wacv} 

\usepackage{multirow}
\usepackage{makecell}
\usepackage[accsupp]{axessibility}  

\definecolor{mydarkblue}{rgb}{0,0.08,1}
\definecolor{mydarkgreen}{rgb}{0.02,0.6,0.02}
\definecolor{myred}{rgb}{1.0,0.0,0.0}
\definecolor{mypurple}{rgb}{0.43,0.0,1.0}

\newcommand{\myheading}[1]{\vspace{2mm}\noindent{\textbf{#1}}}

\def\datasetname{PerMultiSeg}

\def\1n{\mathbf{1}_n}
\def\0{\mathbf{0}}
\def\1{\mathbf{1}}

\def\C{\mathbf{C}}

\def\P{\mathbf{P}}

\def\V{\mathbf{V}}

\definecolor{pink}{rgb}{0.9,0.5,0.5}
\definecolor{purple}{rgb}{0.5, 0.4, 0.8}   
\definecolor{gray}{rgb}{0.3, 0.3, 0.3}
\definecolor{mygreen}{rgb}{0.2, 0.6, 0.2}

\definecolor{greena}{rgb}{0.4, 0.5, 0.1}

\definecolor{bluea}{rgb}{0, 0.4, 0.6}

\definecolor{reda}{rgb}{0.6, 0.2, 0.1}

\newcommand{\cm}[1]{}

\newif\ifshowsolution
\showsolutiontrue

\ifshowsolution

\else

\fi

\definecolor{wacvblue}{rgb}{0.21,0.49,0.74}
\usepackage[pagebackref,breaklinks,colorlinks,allcolors=wacvblue]{hyperref}

\def\wacvPaperID{1289} 
\def\confName{WACV}
\def\confYear{2027}

\title{Where is My Mug? Training-Free Framework for Personalized Image Segmentation}

\author{%
Quang Qui-Vinh Nguyen$^{1,*}$ \quad
Hung Nguyen$^{2,*,\dagger}$ \quad
Rang Nguyen$^{3}$ \quad
Khoi Nguyen$^{1}$\\[2pt]
$^{1}$Qualcomm AI Research$^{\ddagger}$ \quad
$^{2}$National University of Singapore \quad
$^{3}$VinFast, Vietnam\\[2pt]
{\tt\small \{quanguy, khoi\}@qti.qualcomm.com} \quad
{\tt\small hungnm@comp.nus.edu.sg} \quad
{\tt\small rangnhm@gmail.com}
}

\begin{document}
\maketitle

\renewcommand{\thefootnote}{\fnsymbol{footnote}}
\setcounter{footnote}{0}
\footnotetext[1]{Equal contribution.}
\footnotetext[2]{Work done while at Qualcomm.}
\footnotetext[3]{Qualcomm AI Research is an initiative of Qualcomm Technologies, Inc.}
\renewcommand{\thefootnote}{\arabic{footnote}}
\setcounter{footnote}{0}

\begin{figure*}[!t]
    \centering
    \includegraphics[width=.8\linewidth]{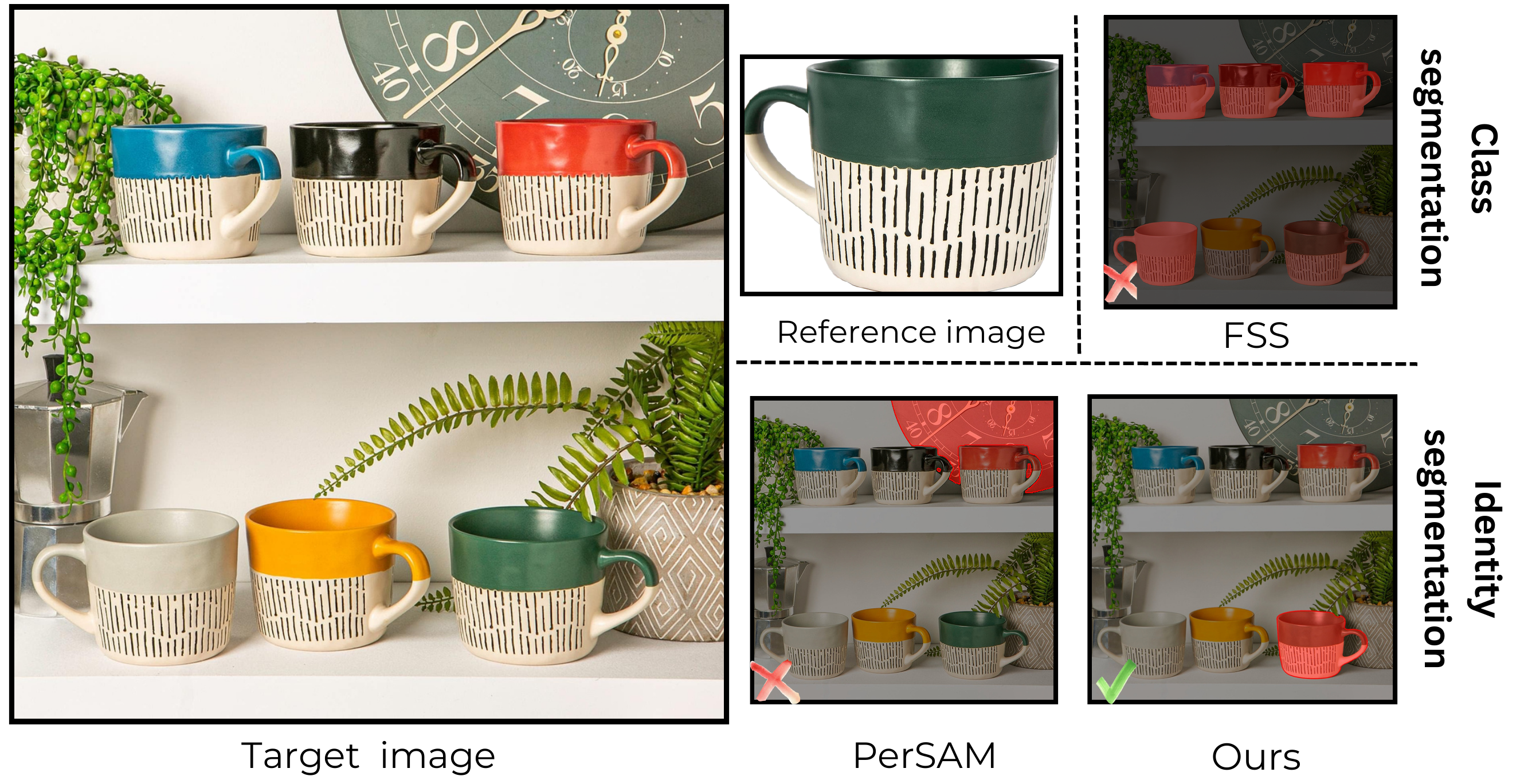}
    \caption{An example from our challenging \datasetname~dataset. Given a distinctive green ceramic mug in the reference image, the task is to accurately identify and segment the same mug among similar ones in the target image. While existing PIS methods such as PerSAM~\cite{persam} fail, our method successfully distinguishes and segments the correct mug. In contrast, few-shot segmentation (FSS) methods would segment all mugs in the scene, without differentiating the specific instance.}
    \label{fig:teaser}
\end{figure*}

\begin{abstract}
\textbf{Personalized Image Segmentation (PIS)} aims to segment the specific object in a target image that matches a user-provided reference. Unlike category-based segmentation or few-shot segmentation, PIS must discriminate between multiple visually similar instances, a requirement central to deployed applications such as augmented reality, personalized content editing, robotics manipulation, inventory tracking, and fine-grained visual retrieval.
We show that the strong numbers reported by existing PIS methods are largely an artifact of single-instance benchmarks: when evaluated on realistic scenes with several same-category look-alikes, prior point-prompt methods collapse, because the correct instance is frequently never surfaced as a prompt. We trace this to a recall bottleneck rather than a mask-quality one, and address it with a training-free, modular system. The pipeline first generates high-recall object proposals using an off-the-shelf class-agnostic detector, then matches each proposal to the reference using foundation-model features (e.g., BLIP-2), and finally refines the selected box with SAM. For robust matching we propose \textbf{top-$K$ Sinkhorn}, a patch-level distance that scores each candidate only on its most confidently matched patches; trimming the Sinkhorn dual potentials recovers the robustness of partial optimal transport at more than $100\times$ lower cost, keeping it inside the per-proposal candidate loop.
To evaluate this regime, we introduce \textbf{\datasetname}, a realistic multi-instance benchmark. Across \datasetname, PerSeg, PerMIS, and DAVIS, our system surpasses PerSAM, $\mathrm{P}^2\mathrm{SAM}$, and PDM by up to $\sim\!48$ mIoU while matching their runtime and using fewer parameters, establishing a strong, deployable training-free baseline for real-world personalized segmentation.
\end{abstract}
\section{Introduction}
\label{sec:intro}

\textbf{Personalized image segmentation (PIS)} aims to segment the specific physical object in a target image that corresponds to a user-provided reference. This capability is increasingly important in deployed perception systems, including warehouse automation, assistive and in-home robotics, augmented reality, and large-scale product retrieval, where models must reliably distinguish one object among several visually similar instances. In these settings, incorrect instance localization directly leads to failure modes such as grasping the wrong item, retrieving the wrong product, or degrading the user experience.

Unlike semantic segmentation, which predicts category-level masks (e.g., ``mug,'' ``chair''), PIS requires instance-identity recognition across changes in viewpoint, lighting, occlusion, and clutter. From a systems perspective, this task stresses the components most essential for real-world reliability: high-recall object proposals, robust instance-level matching, and stable mask generation. Prior works such as PerSAM~\cite{persam} and PDM~\cite{pdm} introduced the concept, but existing benchmarks seldom include multiple same-class objects, limiting their ability to evaluate these critical system behaviors. As illustrated in \cref{fig:teaser}, practical deployments often require selecting a specific green ceramic mug from a group of nearly identical mugs, a task category-based or few-shot segmentation methods struggle to handle.

Although PIS superficially resembles few-shot segmentation (FSS), the two differ fundamentally: FSS generalizes to new semantic \emph{classes} from a few labeled examples \cite{nguyen2019feature,DIaM}, whereas PIS must generalize to new \emph{instances} of the same identity, relying on fine-grained visual cues rather than category-level semantics. This underpins use cases such as instance-specific retrieval~\cite{oquab2023dinov2,blip2,delf}, virtual try-on~\cite{viton,cpvton,vitonhd,stableviton,swifttry,clothformer,fashionvdm}, and robotic manipulation in multi-object scenes~\cite{squeezedet,RISeg,rt-RISeg}, where FSS methods are not suitable replacements.

To meet these demands, we develop a \textbf{training-free, modular system pipeline} that emphasizes robustness and deployment efficiency. Our three-stage design consists of high-recall class-agnostic proposals, foundation-model-based reference--candidate matching, and a mask-generation module. First, class-agnostic detectors ensure that all plausible objects are surfaced; missing the correct instance at this stage is unrecoverable in downstream processing. Second, we perform matching using CLIP~\cite{radford2021learning}, DINOv2~\cite{oquab2023dinov2}, and other foundation-model features. Empirical analysis shows that naive feature averaging is brittle in cluttered scenes, and even patch-level matching with the standard Sinkhorn divergence~\cite{Sejourne2019SinkhornDF} degrades when same-category look-alikes share most of their patch content: high-cost \emph{outlier} patches (occluders, clutter, and non-matching regions) accumulate transport cost and blur the gap between the correct instance and its look-alikes. Robust optimal transport addresses exactly this outlier structure, but solving it directly is too slow to run inside a per-proposal candidate loop. We therefore propose \textbf{top-$K$ Sinkhorn matching}, which scores each candidate only on its most confidently matched patches by discarding the highest-cost patches before summing the transport cost. This recovers the robustness of partial optimal transport~\cite{caffarelli2010partialOT,figalli2010partial,chapel2020partialOT} at more than $100\times$ lower cost (a single sort on top of the standard Sinkhorn pass), making inlier-restricted matching practical for deployment. Finally, the chosen proposal is refined into a segmentation mask. Because each component is modular and swappable without retraining, the system meets diverse hardware and latency budgets.

A major challenge is the lack of datasets that adequately evaluate system-level PIS performance. Existing datasets such as PerSeg~\cite{persam}, PerMIS~\cite{pdm}, and DAVIS adaptations~\cite{davis2017} typically feature only one same-class object per target image. While this leads to high reported mIoU (95--97\%), these benchmarks do not stress the failure modes that dominate real-world performance: proposal recall, instance-level discrimination, and robustness to clutter. Thus, methods that perform well on these datasets often fail in realistic multi-object environments.

To more accurately assess system reliability, we introduce \textbf{\datasetname}, a challenging multi-instance benchmark featuring scenes with 3--4 same-class objects on average, varied viewpoints, natural distractors, and occlusions, better reflecting the operational conditions of robotics, retail perception, and visual search systems.

We evaluate our system across PerSeg, DAVIS, PerMIS, and \datasetname, measuring not only accuracy but also proposal recall, matching reliability, and efficiency trade-offs. Our approach achieves strong improvements, improving on the best prior method by 22.08 mIoU on PerMIS and by over 45 mIoU on \datasetname, demonstrating that a recall-first system, paired with an efficient robust matcher, delivers reliable personalized segmentation in real environments.

Our contributions are:
\begin{itemize}
    \item \textbf{A modular, training-free PIS system} built around a recall-first design: high-recall class-agnostic proposals, foundation-model matching, and SAM refinement. Each module (detector, encoder, matcher) can be swapped independently to meet hardware or latency budgets without retraining, while outperforming prior PIS methods by up to $\sim\!48$ mIoU at comparable runtime.
    \item \textbf{Top-$K$ Sinkhorn matching}, a robust patch-level distance that scores each candidate only on its most confidently matched patches via a single sort on the Sinkhorn dual potentials. To our knowledge it is the first use of inlier-restricted (partial) optimal transport as an instance-level discriminator in personalized segmentation; it recovers the robustness of partial optimal transport at more than $100\times$ lower cost, making inlier-restricted matching practical inside the per-proposal candidate loop.
    \item \textbf{\datasetname}, a realistic multi-instance evaluation benchmark. Motivated by our finding that the high mIoU of existing methods is largely an artifact of single-instance benchmarks, where the dominant failure mode is \emph{proposal recall}, not mask quality, \datasetname\ contains 3--4 same-class objects per target image to stress-test instance-level discrimination that prior PIS benchmarks (PerSeg, PerMIS, DAVIS adaptations) do not capture.
\end{itemize}

\section{Related Work}
\label{sec:related_work}

\myheading{Vision Foundation Models.}
Recent advances in vision foundation models have shown remarkable generalization, making them highly adaptable for diverse tasks. For example, CLIP~\cite{radford2021learning} uses contrastive learning on 400 million image-text pairs for zero-shot visual recognition. BLIP-2~\cite{blip2} combines frozen image encoders and large language models (LLMs) with a lightweight transformer, achieving state-of-the-art results on vision-language tasks. DINOv2~\cite{oquab2023dinov2} learns versatile visual representations via self-supervised learning at both image and patch levels, while SAM~\cite{sam}, pre-trained on one billion masks, enables promptable and highly accurate segmentation. Despite their versatility, these models struggle with personalized image segmentation, as they are not trained to segment the same object across images. In this paper, we propose a \textbf{training-free framework} that leverages off-the-shelf vision foundation models for one-shot personalized segmentation, enabling precise localization and segmentation of uniquely identified objects.

\myheading{Image Segmentation.}
Image segmentation has progressed significantly across semantic, instance, panoptic, and few-shot segmentation tasks~\cite{long2015fully,strudel2021segmenter,carion2020end,cheng2021per,li2022exploring,nguyen2019feature}, supported by advances in object proposal generation. However, most traditional approaches rely on extensive task-specific training and fine-tuning, limiting their ability to adapt to new scenarios without retraining.
Recent class-agnostic segmentation models, such as SAM~\cite{sam}, achieve broad generalization through large-scale pretraining and prompt-based inference. Painter~\cite{wang2023images} frames vision tasks as image-to-image prediction using masked image modeling, while SegGPT~\cite{wang2023seggpt} improves generalization via random colorization. SEEM~\cite{zou2023segment} further expands generality by supporting spatial, visual, and text prompts within a single framework.
Despite their versatility, these models are not designed for object-specific segmentation, where the objective is to identify and segment the exact same object instance across images. This fine-grained identity matching remains a key gap in existing foundation-model-based segmentation systems, one that our work directly addresses.

\myheading{Personalized Image Segmentation.}
Prior works have explored personalized segmentation, but existing approaches struggle to robustly segment the same object instance with its unique identity. PerSAM~\cite{persam} adapts SAM for personalized and video segmentation using either zero-shot inference or minimal fine-tuning, extracting features from reference and target images to produce confidence maps. $\text{P}^2\text{SAM}$~\cite{p2sam} improves robustness by filtering outlier prompts, while Matcher~\cite{matcher} combines class-agnostic segmentation with bidirectional matching for dense correspondence. PDM~\cite{pdm} instead leverages intermediate features from pre-trained text-to-image models to compute appearance and semantic similarity. These methods are largely evaluated on benchmarks with only a single same-category object per target image, which does not reflect real deployment settings.

\myheading{Optimal Transport for Feature Matching.}
OT has been applied across vision tasks: DeepEMD~\cite{deepemd} uses earth-mover's distance on patch features for few-shot classification; OTAM~\cite{otam} uses temporal alignment for few-shot video recognition; SuperGlue~\cite{superglue} and LoFTR~\cite{loftr} employ Sinkhorn and dual-softmax to obtain soft keypoint assignments. On the theoretical side, partial and robust OT relax marginal constraints to tolerate outliers~\cite{caffarelli2010partialOT,figalli2010partial,chapel2020partialOT,mukherjee2021outlierOT,balaji2020robustOT}.
Despite this literature, robust and partial OT have seen little use as an instance-level discriminator for personalized segmentation, where the matcher must run inside a per-proposal candidate loop. The closest work, Matcher~\cite{matcher}, uses full balanced OT to score mask quality, not partial OT to discriminate a specific instance. Existing methods either aggregate over all pairs (DeepEMD, full Sinkhorn) or trim correspondences post-hoc (SuperGlue); Screenkhorn~\cite{alaya2019screenkhorn} screens dual variables before solving to accelerate exact OT. In contrast, our top-$K$ Sinkhorn trims the transport objective itself, recovering partial-OT-level robustness at a fraction of the cost. To our knowledge this is the first use of inlier-restricted OT as an instance-level discriminator in personalized segmentation.
\section{Our \datasetname~Dataset}
\label{sec:dataset}

\begin{figure*}[!t]
  \centering
  \includegraphics[width=\linewidth]{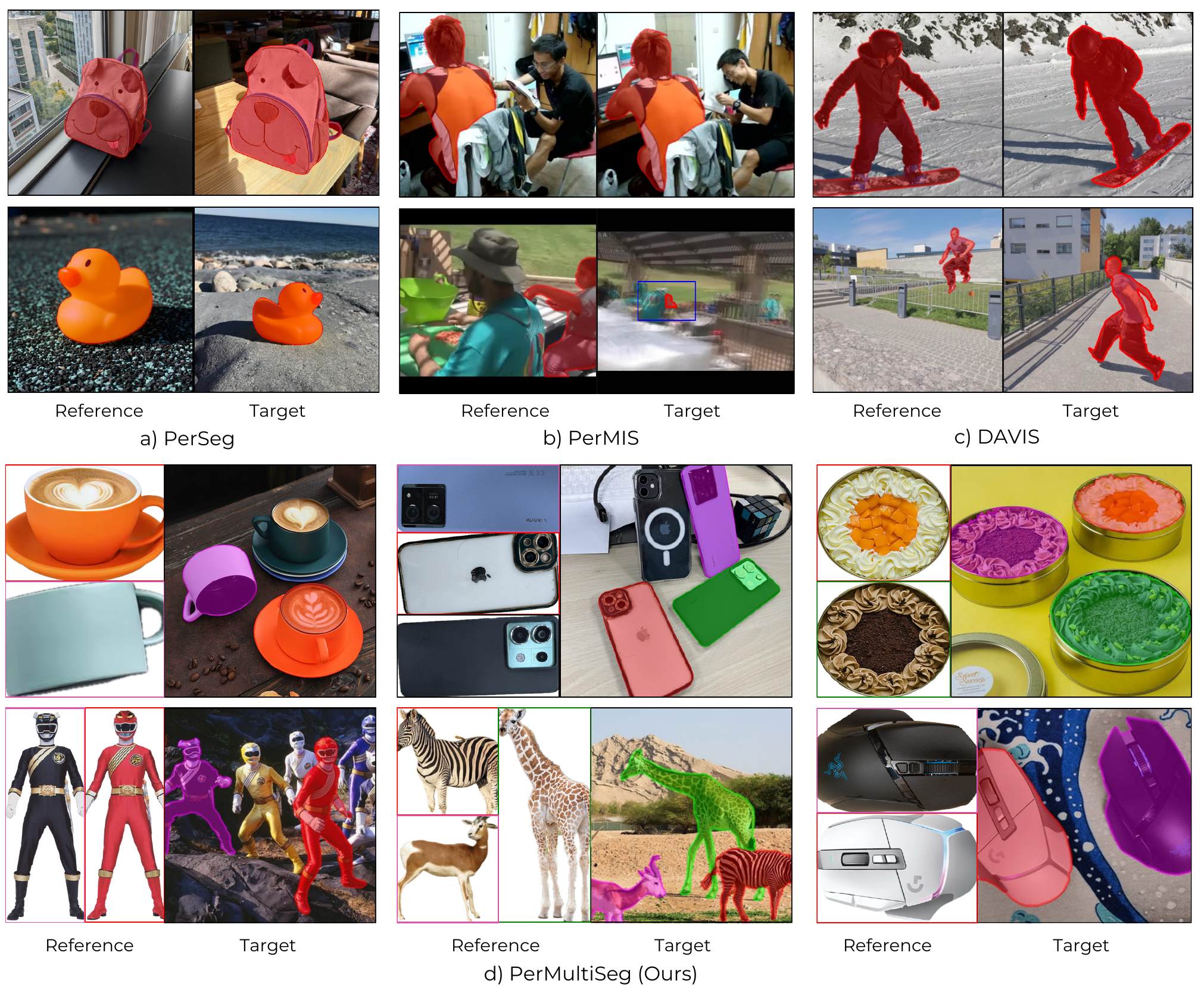}
     \caption{Examples of existing PIS benchmarks (left: reference, right: target). PerSeg has only one instance per target image; PerMIS~\cite{pdm} suffers from low-quality and near-duplicate frames (blue rectangle: targets too small to identify); DAVIS~\cite{davis2017} exhibits minimal frame-to-frame variation. \datasetname\ provides multiple same-category instances per target image (e.g., three mugs, three smartphones).}
   \label{fig:dataset_compare}
\end{figure*}

\begin{table}[t]
\centering
\small
\setlength{\tabcolsep}{4pt}
\caption{Comparison of \datasetname~with prior PIS datasets.}
\resizebox{\linewidth}{!}{
\begin{tabular}{l c c c c}
\toprule
\textbf{Dataset} & \textbf{\# Ref} & \textbf{\# Identities} & \textbf{\# Categories} & \textbf{\# Obj/Img}  \\
\midrule
PerSeg~\cite{persam}  &  40 & 40 & 40 & 1.00 \\
PerMIS~\cite{pdm}  &  140 & 140 & 16 & 3.10  \\
DAVIS~\cite{davis2017}  &  91 & 91 &  30 & 3.18  \\
\midrule
\textbf{\datasetname}  & \textbf{2,147} & \textbf{1,377}& \textbf{45} & \textbf{3.67}  \\
\bottomrule
\end{tabular}
}
\label{tab:comparison}
\end{table}
\vspace{-8pt}

Existing PIS benchmarks are limited along two axes (\cref{fig:dataset_compare}, \cref{tab:comparison}): \emph{(i)} they rarely contain multiple same-category instances in a single target: PerSeg~\cite{persam} is essentially single-instance, while PerMIS~\cite{pdm} and DAVIS~\cite{davis2017} are video-derived with minimal reference-to-target variation; \emph{(ii)} they cover fewer than 100 categories at small scale. As stress tests for fine-grained instance discrimination, these benchmarks fall short.

To address these gaps, we introduce \datasetname, a larger and more challenging PIS benchmark. Each target image contains multiple same-category objects ($3.67$ per image on average) paired with multiple references, and the dataset spans 45 categories collected across diverse environments, lighting conditions, and viewpoints. In total, \datasetname\ provides $1{,}079$ target images, $2{,}147$ reference images, and $3{,}964$ annotated instances. Each reference is manually matched to its counterpart in the target and annotated with a precise binary mask. Crucially, the distractors are selected to be the \emph{same product and same colour} as the reference, that is, visually near-identical units that differ only in physical identity, pose, and wear; this removes category- and colour-level shortcuts and forces the matcher onto fine-grained localised evidence. Because the criterion needs near-identical units to co-occur naturally, categories skew toward portable manufactured goods and scenes toward indoor settings, which bounds external validity but not the comparisons, since all methods see identical pairs. Detailed dataset statistics, the collection and annotation protocol, selection criterion, and category-level breakdowns (Figs.~\ref{fig:dataset-analysis-suppl}) are provided in the supp.

\section{Our Framework}
\label{sec:method}

\begin{figure*}[t]
  \centering
  \includegraphics[width=1\linewidth]{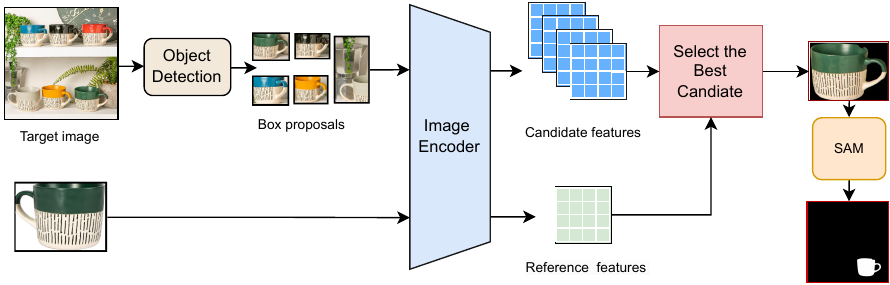}
       \caption{Given a reference and target image, we first use an object detector to identify all potential object candidates in the target image. These candidates, along with the reference image, are then passed through an image encoder to extract feature representations. Next, we determine the best-matching candidate by computing the distance between each candidate's features and the reference features. Finally, the selected candidate's bounding box is fed into SAM, which generates the final segmentation mask.}
   \label{fig:framework}
\end{figure*}

\begin{table}[t]
\small
\centering
\setlength{\tabcolsep}{3pt}
\caption{Proposal-stage recall (before matching) and end-to-end inference time on \datasetname. Each method is evaluated under its native protocol (details in supp.). Our method achieves significantly higher recall, improving final segmentation.}
\resizebox{\linewidth}{!}{
\begin{tabular}{l c c c c}
  \toprule
  & \textbf{PerSAM}~\cite{persam} & \textbf{PDM}~\cite{pdm} & \textbf{$\text{P}^2\text{SAM}$}~\cite{p2sam} & \textbf{Ours} \\
  \midrule
  \textbf{Recall (\%) $\uparrow$} & 34.33 & 30.60 & 34.33 & 89.55 \\
  \textbf{Inf. time (s) $\downarrow$} & 1.07 & 26.29 & 1.44 & 1.02 \\
  \bottomrule
\end{tabular}
}
\label{tab:analysis-results}
\end{table}
\subsection{Overall System}

In multi-object scenes, failures in personalized segmentation arise from missed localization, not mask refinement: once the correct instance is absent from the proposal set, no downstream step recovers it. This is the bottleneck in point-prompt pipelines, whose SAM confidence maps miss the correct instance in clutter, so PerSAM~\cite{persam} and PDM~\cite{pdm} degrade sharply on \datasetname\ despite strong mask generators (low proposal recall, \cref{tab:analysis-results}). We therefore adopt a recall-first, proposal-based framework~\cite{maskrcnn} that decouples candidate generation from instance matching, in three~stages:
\begin{enumerate}
    \item \textbf{Class-agnostic object proposals}: detect bounding boxes for all potential objects without using category labels (\cref{sec:mask_object_proposal});
    \item \textbf{Reference--candidate matching}: identify the proposal that best corresponds to the reference object (\cref{sec:candidate_selection});
    \item \textbf{Mask refinement}: prompt SAM with the selected box.
\end{enumerate}
Decoupling these stages keeps each module replaceable under different latency or hardware budgets without retraining. The pipeline is illustrated in \cref{fig:framework}; despite being training-free, it matches the runtime of PerSAM and $\text{P}^2\text{SAM}$ while substantially outperforming them (\cref{tab:analysis-results}).

\subsection{Class-agnostic Object Proposals}
\label{sec:mask_object_proposal}
We capture a comprehensive set of object proposals without considering their classes, leveraging any object detector such as YOLO-World~\cite{yoloword} or CoDet~\cite{ma2023codet}, both trained with the open vocabulary of Object365~\cite{objects365} (\cref{tab:ablate-proposal}). After this step, we obtain $B$ class-agnostic box proposals $\{b_i\}_{i=1}^B$, which achieve substantially higher proposal recall than prior PIS methods (\cref{tab:analysis-results}).

\subsection{Reference--Candidate Matching}
\label{sec:candidate_selection}
Given the proposals, we extract features for each $b_i$ and the reference $I_r$ using any vision foundation encoder~\cite{oquab2023dinov2,radford2021learning,blip,sam,sam2} (compared in \cref{tab:ablate-image_enc}). When a $[\mathrm{CLS}]$ token is available we take it as the global feature $\mathbf{v}_0\in\mathbb{R}^d$ and the remaining $N=HW$ patch tokens as $\V\in\mathbb{R}^{N\times d}$; the global distance $d_\mathrm{g}(\mathbf{v}_0^{r},\mathbf{v}_0^{(i)})$ is cosine distance.

\myheading{Patch-level distance: limitations of full Sinkhorn.}
A natural patch-level distance is the entropic-regularised Sinkhorn divergence~\cite{Sejourne2019SinkhornDF}
\begin{equation}
\label{eq:sinkhorn_div}
\begin{aligned}
\mathcal{S}_\epsilon(\V^{r}, \V^{(i)})
&= \mathrm{OT}_\epsilon(\V^{r}, \V^{(i)}) \\
&\quad - \tfrac{1}{2}\mathrm{OT}_\epsilon(\V^{r}, \V^{r})
- \tfrac{1}{2}\mathrm{OT}_\epsilon(\V^{(i)}, \V^{(i)}),
\end{aligned}
\end{equation}
where $\mathrm{OT}_\epsilon$ is the entropic OT cost~\cite{cuturi2013sinkhorn} (full definition in supp.). In practice, we use only the cross term $\mathrm{OT}_\epsilon(\V^r,\V^{(i)})$, since the de-biased self-terms add no discriminative signal for candidate ranking (verified empirically in supp., \cref{tab:selfterm}). However, even the cross term alone aggregates over \emph{all} $N\times N$ patch pairs. In multi-instance PIS this is harmful: a large fraction of patches in $\V^{r}$ and $\V^{(i)}$ correspond to background, partial occluders, or shared category structure common to all same-category candidates. These high-cost outlier patches accumulate transport cost roughly in proportion to their number, compressing the gap between the correct instance and its look-alikes. The result is that wrong candidates and the correct one yield comparable $\mathcal{S}_\epsilon$ scores.

\myheading{Top-$K$ Sinkhorn matching.}
\label{sec:topk}
We exploit the inlier structure of the problem and score each candidate only on its most confidently matched patches. Let $\mathbf f^{\star},\mathbf g^{\star}\in\mathbb{R}^{N}$ be the Sinkhorn dual potentials of the cross term $\mathrm{OT}_\epsilon(\mathbf{V}^{r}, \mathbf{V}^{(i)})$ and $\mathbf{a},\mathbf{b}$ the (uniform) marginal weights. By Sinkhorn duality~\cite{cuturi2013sinkhorn},
\begin{equation}
    \mathrm{OT}_\epsilon(\V^{r}, \V^{(i)})
    = \sum_{j=1}^{N} a_j\,f^{\star}_j + \sum_{k=1}^{N} b_k\,g^{\star}_k,
    \label{eq:duality}
\end{equation}
so each per-patch term $a_j f^{\star}_j$ is the marginal cost of transporting source patch $j$: it is \emph{small} (often negative) for patches with a close match in $\V^{(i)}$ and \emph{large} for patches without, exactly the inlier/outlier structure our matcher relies on.

Given a \emph{keep-ratio} $K\in(0,1]$ (the fraction of patches we retain on each side, \emph{not} an integer count, so that $K=1/3$ keeps the cleanest one-third), let $M=\lfloor K N\rfloor$ and let $(\mathbf{a}\!\odot\!\mathbf f^{\star})_{(1)}\le\dots\le(\mathbf{a}\!\odot\!\mathbf f^{\star})_{(N)}$ be the per-patch contributions sorted ascending (here $(\cdot)_{(m)}$ denotes the $m$-th order statistic, i.e.\ the $m$-th smallest element). The top-$K$ Sinkhorn distance keeps only the smallest $M$ contributions on each~side:
\begin{equation}
    \widetilde{\mathrm{OT}}^{\,K}_\epsilon\!\big(\mathbf{V}^{r},\mathbf{V}^{(i)}\big)
    = \!\!\sum_{m=1}^{M}\!(\mathbf{a}\!\odot\!\mathbf f^{\star})_{(m)}
    + \!\!\sum_{m=1}^{M}\!(\mathbf{b}\!\odot\!\mathbf g^{\star})_{(m)}.
    \label{eq:topk_ot}
\end{equation}
At $K=1$, \cref{eq:topk_ot} recovers full entropic OT; smaller $K$ progressively restricts the score to the cleanest inlier subset. This is motivated by partial optimal transport~\cite{caffarelli2010partialOT,figalli2010partial,chapel2020partialOT}, which down-weights outlier mass to gain robustness; rather than solving the partial-OT program directly (which modifies the Sinkhorn iteration and is too slow for the per-proposal candidate loop), we trim the dual potentials of the \emph{full} solve as a cheap surrogate, and verify empirically (\cref{tab:ablate-distance-method}) that it recovers partial-OT-level accuracy. We adopt
\begin{equation}
    d_\mathrm{p}(\V^{r},\V^{(i)}) = \widetilde{\mathrm{OT}}^{\,K}_\epsilon\!\big(\V^{r},\V^{(i)}\big),
    \label{eq:patch_dist}
\end{equation}
with $K=1/3$ throughout (trim-ratio sweep in supp., \cref{tab:ablate-alpha}). Trimming is applied directly to the dual potentials returned by the Sinkhorn solver, so top-$K$ Sinkhorn requires only an $\mathcal{O}(N\log N)$ sort per pair on top of the standard Sinkhorn forward pass; no modification of the iteration is needed. In practice this makes robust inlier-restricted matching more than $100\times$ cheaper than a direct partial-OT solver (\cref{tab:ablate-distance-method}), cheap enough to remain inside the candidate loop at deployment.

\myheading{Candidate selection and segmentation.}
The best candidate combines patch- and global-level distances with weights $\alpha_\mathrm{p},\alpha_\mathrm{g}$:
\begin{equation}
    i^{\star} = \arg\min_{i}\;\big[\alpha_\mathrm{p}\,d_\mathrm{p}(\V^{r},\V^{(i)}) + \alpha_\mathrm{g}\,d_\mathrm{g}(\mathbf{v}_0^{r},\mathbf{v}_0^{(i)})\big].
    \label{equation:combine-mask}
\end{equation}
The selected box $b_{i^{\star}}$ then prompts SAM~\cite{sam} to produce the final segmentation mask.

\section{Experiments}

\subsection{Experimental Setup}

\myheading{Datasets.}
We evaluate on PerSeg~\cite{persam}, DAVIS~\cite{davis2017}, PerMIS~\cite{pdm}, and our proposed \datasetname~(\cref{sec:dataset}). Component ablations use our validation set (135 reference--target pairs). For VOS, DAVIS uses the first frame as reference and subsequent frames as targets.

\myheading{Metrics.}
Following~\cite{persam,pdm}, we report mIoU and bIoU. On \datasetname\ we additionally report Recall (detailed in supp.). For VOS we report $\mathcal{J}$, $\mathcal{F}$, and $\mathcal{J\&F}$.

\myheading{Implementation details.}
All experiments use an A100 40GB GPU with PyTorch 2.6. Default configuration: BLIP-2~\cite{blip2} (ViT-G/14~\cite{EVA}) for feature extraction, YOLOv11 for detection (confidence $10^{-3}$), SAM~\cite{sam} (ViT-H) for mask generation. Weights: $\alpha_\mathrm{p}=0.75$, $\alpha_\mathrm{g}=0.25$. We solve the entropic OT via GeomLoss~\cite{feydy2019sinkhorn} and apply the top-$K$ Sinkhorn matching of \cref{sec:topk} with $K=1/3$. Following E5 (supp.), we use only the cross transport term, since the de-biased self-terms add no accuracy at $3\times$ the cost.

\myheading{Baselines.} PerSAM~\cite{persam}, PDM~\cite{pdm}, $\text{P}^2\text{SAM}$~\cite{p2sam}, all at default settings with the same SAM checkpoint. For $\text{P}^2\text{SAM}$ we report its \emph{official} default configuration (supp., \cref{sec:baseline_config}).

\subsection{Comparisons with Prior Approaches}
\vspace{-3pt}
\begin{table*}[t]
\centering
\setlength{\tabcolsep}{6pt}
\caption{Comparison on the {\datasetname} dataset. Recall is measured at the final-prediction stage: a method is counted as a hit if its output mask has IoU $>$ 0.5 with the ground-truth mask (applied uniformly to all methods, regardless of prompt type). ``Box Prompt'' refers to using bounding box inputs for SAM, while ``Point Prompt'' denotes using point-based inputs.}
\vspace{-6pt}
\begin{tabular}{l l l c c c c}
  \toprule
  {\textbf{Method}} & {\textbf{Prompt}} & {\textbf{Backbone}} & \textbf{mIoU $\uparrow$} & \textbf{bIoU $\uparrow$} & \textbf{Recall $\uparrow$} & \textbf{Params $\downarrow$} \\
  \midrule
  PerSAM~\cite{persam} & Point & SAM & 21.65 & 14.82 & 25.29 & 1.9B \\
  PDM~\cite{pdm} & Point & SDXL-turbo & 23.65 & 12.82 & 43.91 & 2.6B \\
  P$^2$SAM~\cite{p2sam} & Point & SAM & 31.58 & 27.90 & 24.57 & 1.9B \\
  \midrule
  SAM2$^\ddagger$~\cite{sam2} & 1-shot & SAM2 & 38.52 & 33.10 & 41.20 & 0.6B \\
  SAM3$^\ddagger$~\cite{sam3} & Concept & SAM3 & 44.73 & 40.95 & 48.15 & 2.1B \\
  SAM3 + Qwen2.5-VL$^\ddagger$~\cite{sam3,qwen25vl} & Caption & SAM3+Qwen & 70.11 & 64.57 & 71.09 & 9.1B \\
  \midrule
  Ours (SAM detector) & Box & SAM & 74.33 & 68.31 & 76.78 & \textbf{1.6B} \\
  Ours (YOLOv11 detector) & Box & SAM & \textbf{80.07} & \textbf{75.15} & \textbf{82.29} & \textbf{1.6B} \\
  \bottomrule
\end{tabular}
\\
{\scriptsize $^\ddagger$Address a different task (VOS/PCS: segment \emph{all} matching instances, not one specific instance); shown for context, see text.}
\label{tab:permultiseg-results}
\end{table*}

\vspace{-8pt}

\begin{table}[t]
\small
\centering
\setlength{\tabcolsep}{3pt}
\caption{Comparison on PerSeg, DAVIS, and PerMIS. PerSAM-F uses 10\,s test-time fine-tuning per target.}
\vspace{-6pt}
\resizebox{\linewidth}{!}{
\begin{tabular}{l c ccc c}
  \toprule
  {\textbf{Method}} &
  \multicolumn{1}{c}{\textbf{PerSeg}} &
  \multicolumn{3}{c}{\textbf{DAVIS2017}} &
  \multicolumn{1}{c}{\textbf{PerMIS}} \\
  \cmidrule(lr){2-2} \cmidrule(lr){3-5} \cmidrule(lr){6-6}
   & \textbf{mIoU $\uparrow$} &
   \textbf{J\&F $\uparrow$} & \textbf{J $\uparrow$} & \textbf{F $\uparrow$} &
   \textbf{mIoU $\uparrow$} \\
  \midrule
  PerSAM~\cite{persam} & 89.30 & 66.90 & 71.30 & 75.10 & - \\
  PerSAM-F$^\dagger$~\cite{persam} & 95.30 & 76.10 & 74.90 & 79.70 & 16.50 \\
  Matcher~\cite{matcher} & - & 73.50 & 70.00 & 77.50 & - \\
  PDM~\cite{pdm} & 95.40 & 75.80 & 72.90 & 80.10 & 42.30 \\
  $\text{P}^2\text{SAM}$~\cite{p2sam} & 95.70 & - & - & - & 43.57 \\
  \midrule
  Ours & \textbf{95.75} & \textbf{78.68} & \textbf{76.44} & \textbf{80.92} & \textbf{65.65} \\
  \bottomrule
\end{tabular}
}
\vspace{-8pt}

\label{tab:multi-dataset-results}
\end{table}

\myheading{Quantitative results.}
\cref{tab:permultiseg-results} shows our approach exceeds $\text{P}^2\text{SAM}$ by more than 45 mIoU on \datasetname, confirming that existing methods fail in multi-instance scenes. Even the detector-free variant (SAM AMG) surpasses all baselines by a large margin. SAM2 and SAM3 target related but different tasks; augmenting SAM3 with a Qwen2.5-VL caption closes much of the gap but still falls 10 mIoU short, confirming that PIS requires visual instance matching that text descriptions cannot replace. On PerMIS and DAVIS (\cref{tab:multi-dataset-results}), our method matches or exceeds all baselines.

\begin{figure*}[t]
  \centering
  \includegraphics[width=0.87\linewidth]{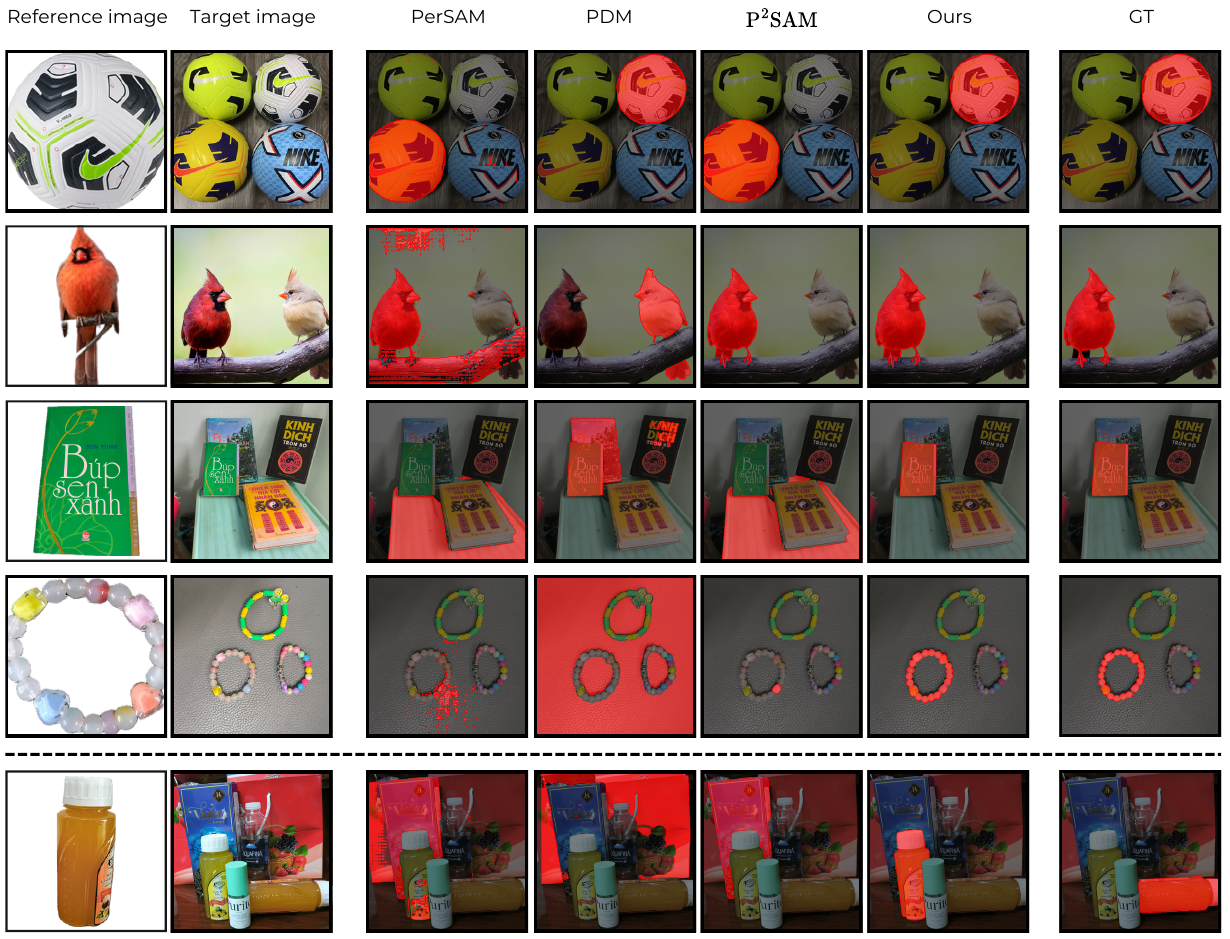}
     \caption{Qualitative results on \datasetname\ (rows 1--4: success; row 5: failure).}
   \label{fig:qualitative_pms}
\end{figure*}

\myheading{Qualitative results.}
\cref{fig:qualitative_pms} shows representative results on \datasetname. In successful cases (rows 1--4) the correct instance is precisely segmented despite multiple same-category distractors. The failure case (last row) occurs when the correct instance has its discriminative label hidden while a distractor displays a similar label facing the camera; any appearance-based matcher will prefer the more visible match. Supp.\ \cref{fig:hardcases_grid} gives further examples; \cref{tab:error_decomposition} reports a stage-wise error decomposition.

\vspace{-4pt}
\myheading{On fair comparison.}
Our pipeline uses more pretrained components than the baselines, and uses box rather than point prompts; three controls show the gains are not solely explained by either. \emph{(i)} \textbf{At an identical prompt type}, our pipeline with a center-point prompt (supp., \cref{tab:few-shots-method}) reaches $55.41$ mIoU against $31.58$ for $\text{P}^2\text{SAM}$, a $+23.83$ gap under the interface the baselines use, roughly half of the total $+48.49$. \emph{(ii)} The \textbf{Ours (SAM detector)} row uses \emph{only} SAM (same backbone as PerSAM, no external detector) and still achieves $+42$ mIoU over $\text{P}^2\text{SAM}$. \emph{(iii)} With the encoder fixed, switching from a single global vector to patch-level OT yields large gains (e.g.\ $+13$ mIoU on DINOv2, CLS$\to$patch-level OT), and top-$K$ Sinkhorn adds a further $+2.9$ mIoU over full Sinkhorn on BLIP-2 (\cref{tab:ablate-distance-method}).

\subsection{Validating the Matcher Choice}
\label{sec:ablation_topk}

All rows in \cref{tab:ablate-distance-method} share the same frozen proposals, SAM checkpoint, and BLIP-2 encoder, differing only in the patch distance $d_\mathrm{p}$, on the 135-pair \emph{validation} split; the deployed last row scores $80.07$ on test.

\begin{table}[t]
\small
\centering
\caption{Matcher ablation on the \datasetname\ \emph{validation} split (135 pairs). All rows use BLIP-2 + $\alpha_\mathrm{p}=0.75$ + the same frozen proposals (paper default). Top-$K$ Sinkhorn is indistinguishable from the partial-OT solver at this sample size, at a fraction of its latency.}
\vspace{-4pt}
\resizebox{\linewidth}{!}{
\begin{tabular}{l c c c c}
\toprule
\textbf{Matcher} & \textbf{mIoU $\uparrow$} & \textbf{Recall $\uparrow$} & \textbf{bIoU $\uparrow$} & \textbf{ms/pair $\downarrow$} \\
\midrule
Avg.\ feature map & 65.29 & 66.44 & 60.97 & 0.5 \\
{[CLS]} token only & 83.86 & 85.58 & 78.77 & 0.07 \\
\midrule
Full Sinkhorn ($K{=}1$) & 85.04 & 86.83 & 79.93 & 8 \\
Sinkhorn + mutual-NN filter & 81.70 & 82.92 & 76.92 & 10 \\
Partial OT ($m{=}1/3$)~\cite{chapel2020partialOT} & 87.97 & 89.83 & 82.69 & 848 \\
\midrule
\textbf{Top-$K$ Sinkhorn ($K{=}1/3$)} & \textbf{87.97} & \textbf{89.83} & \textbf{82.69} & \textbf{8} \\
\bottomrule
\end{tabular}
}
\label{tab:ablate-distance-method}
\end{table}

\myheading{Robust matching at deployable cost.}
Top-$K$ Sinkhorn selects the same candidate as the principled partial-OT solver~\cite{chapel2020partialOT} on all 135 pairs, so their masks are identical, at \textbf{more than $100\times$ lower latency} (8 vs.\ 848\,ms per pair; supp., \cref{sec:matcher_stats}). Against full Sinkhorn it gains $+2.93$ mIoU at the shared default $\alpha_\mathrm{p}{=}0.75$ and $+1.74$ best-versus-best (supp., \cref{tab:alpha_sweep}). Matching is only $\sim$1.5\% of end-to-end latency (supp., \cref{tab:latency-breakdown}), effectively free at deployment.

\myheading{Simpler matchers are insufficient.}
Single-vector matching is brittle: mean-pooling and even the [CLS] token discard the spatial structure this task needs. Full Sinkhorn is diluted by the $\sim$2/3 of patches that are background or shared-category content; trimming the \emph{correspondences} post-hoc (mutual-NN, SuperGlue-style) is worse, breaking the transport coupling. Top-$K$ instead trims the per-patch dual contributions, keeping only confident matches; the trim is robust for any $K\in[1/3,1/2]$, and we use the most aggressive, $K{=}1/3$.

\subsection{Component Ablations}

\myheading{Object proposals.} We compare class-agnostic detectors in the supp (\cref{tab:ablate-proposal}) and sweep the YOLOv11 confidence threshold (\cref{tab:threshold}). SAM's Automatic Mask Generator achieves the highest recall on \datasetname\ (99.38\%) but at roughly 25$\times$ the latency of YOLOv11; YOLOv11 offers the best recall--accuracy--size trade-off across both datasets and is our default at $\tau{=}10^{-3}$, where lower thresholds monotonically improve final mIoU because top-$K$ matching absorbs the extra outlier proposals.

\myheading{Image encoder.} \cref{tab:ablate-image_enc} compares six encoders under three matchers. Two findings: (1) patch-level OT massively outperforms CLS-only for self-supervised encoders (DINOv2: $+$13 mIoU); (2) top-$K$ Sinkhorn improves over full Sinkhorn for \emph{every} encoder tested (per-encoder sweep in supp., \cref{fig:ksweep_encoders}), confirming inlier-restricted scoring as a general matching primitive rather than a BLIP-2-specific trick. BLIP-2 delivers the highest absolute accuracy (87.97) and is our default.

\begin{table}[t]
\centering
\small
\caption{Study on image encoder. The Top-$K$ column reports the best result per encoder from the $K$ sweep (\cref{fig:ksweep_encoders}, supp.); $\Delta$ is vs.\ CLS-only.}
\vspace{-4pt}
\resizebox{\linewidth}{!}{
\begin{tabular}{l c c c c}
\toprule
\textbf{Encoder} & \textbf{CLS only} & \textbf{Mean pool} & \textbf{Top-$K$} & \textbf{$\Delta$ vs CLS} \\
\midrule
BLIP-2~\cite{blip2}       & 83.86 & 83.16 & \textbf{87.97} & +4.11 \\
SigLIP-2~\cite{siglip2}   & 77.02 & 84.54 & \textbf{86.50} & +9.48 \\
EVA-02 CLIP~\cite{EVA}    & 71.67 & 72.32 & \textbf{75.69} & +4.02 \\
DINOv2~\cite{oquab2023dinov2} & 66.96 & 70.92 & \textbf{80.71} & +13.75 \\
DINOv3~\cite{dinov3}    & 65.21 & 70.94 & \textbf{78.26} & +13.05 \\
CLIP~\cite{radford2021learning} & 71.35 & 74.76 & \textbf{74.82} & +3.47 \\
\bottomrule
\end{tabular}
}
\label{tab:ablate-image_enc}
\vspace{-12pt}
\end{table}
\vspace{-4pt}

\section{Conclusion}
\vspace{-4pt}

We showed that strong reported PIS numbers are largely a single-instance artifact; proposal recall is the real bottleneck. Our training-free, recall-first system introduces \textbf{top-$K$ Sinkhorn}, matching partial-OT accuracy at more than $100\times$ lower cost. Across \datasetname\ and three existing datasets, it surpasses prior methods by up to $\sim\!48$ mIoU at comparable runtime.


{
    \small
    \bibliographystyle{ieeenat_fullname}
    \bibliography{main}
}

\clearpage
\appendix
\clearpage
\setcounter{page}{1}
\maketitlesupplementary

\section{Optimal Transport Background}
\label{sec:ot_background}

This section provides the formal definitions of entropic regularised optimal transport and Sinkhorn divergence used in \cref{sec:candidate_selection} of the main paper.

\myheading{Entropic regularised OT.}
Let $\alpha = \tfrac{1}{N}\sum_{j=1}^{N}\delta_{\V^{r}_{j}}$ and $\beta = \tfrac{1}{N}\sum_{k=1}^{N}\delta_{\V^{(i)}_{k}}$ be the empirical measures over the reference and candidate patch grids, with uniform marginal weights $\mathbf{a},\mathbf{b}\in\mathbb{R}^{N}_{+}$. Given a ground cost matrix $\C\in\mathbb{R}_{+}^{N\times N}$ whose entry $\C_{j,k}$ is the squared Euclidean distance between $\V^{r}_{j}$ and $\V^{(i)}_{k}$, the entropic regularised OT cost~\cite{cuturi2013sinkhorn} is
\begin{equation}
\mathrm{OT}_\epsilon(\alpha, \beta)
= \min_{\P \in U(\mathbf{a},\mathbf{b})} \sum_{j,k} \P_{j,k}\,\C_{j,k}
+ \epsilon \sum_{j,k} \P_{j,k}\log \P_{j,k},
\label{eq:OT_suppl}
\end{equation}
where $\P\in[0,1]^{N\times N}$ is the transport plan, $U(\mathbf{a},\mathbf{b})$ is the polytope of joint distributions with marginals $\mathbf{a},\mathbf{b}$, and $\epsilon>0$ is the entropic regularisation strength. We solve \cref{eq:OT_suppl} with the Sinkhorn algorithm via the GeomLoss implementation~\cite{feydy2019sinkhorn}, which returns the dual potentials $\mathbf f^{\star},\mathbf g^{\star}\in\mathbb{R}^{N}$ satisfying $\mathrm{OT}_\epsilon(\alpha,\beta) = \sum_j a_j f^{\star}_j + \sum_k b_k g^{\star}_k$.

\myheading{Sinkhorn divergence.}
\Cref{eq:sinkhorn_div} of the main paper defines the de-biased Sinkhorn divergence~\cite{Sejourne2019SinkhornDF}
\begin{align}
\mathcal{S}_\epsilon(\V^{r}, \V^{(i)})
&= \mathrm{OT}_\epsilon(\V^{r}, \V^{(i)}) \notag\\
&\quad - \tfrac{1}{2}\mathrm{OT}_\epsilon(\V^{r}, \V^{r}) - \tfrac{1}{2}\mathrm{OT}_\epsilon(\V^{(i)}, \V^{(i)}),
\label{eq:sinkhorn_div_suppl}
\end{align}
which subtracts the two self-transport terms to remove the entropic bias of $\mathrm{OT}_\epsilon$. To compute $\mathcal{S}_\epsilon$, we solve three independent $\mathrm{OT}_\epsilon$ subproblems with cost matrices
$\C^{r,i}_{j,k}=\lVert\V^{r}_j - \V^{(i)}_k\rVert_2$,
$\C^{r,r}_{j,k}=\lVert\V^{r}_j - \V^{r}_k\rVert_2$, and
$\C^{i,i}_{j,k}=\lVert\V^{(i)}_j - \V^{(i)}_k\rVert_2$,
and combine them via \cref{eq:sinkhorn_div_suppl}.

\myheading{Top-$K$ Sinkhorn and partial OT.}
\Cref{eq:topk_ot} of the main paper trims the dual sum of the cross term $\mathrm{OT}_\epsilon(\V^{r},\V^{(i)})$ to its $\lfloor KN\rfloor$ smallest per-patch contributions on each side. This is motivated by the partial optimal transport problem~\cite{caffarelli2010partialOT,figalli2010partial,chapel2020partialOT}, in which a fraction $K\in(0,1]$ of mass on each side is allowed to be ignored:
\begin{equation*}
\inf_{\substack{\alpha'\le\alpha,\;\beta'\le\beta\\ |\alpha'|=|\beta'|=K}}\;\mathrm{OT}_\epsilon(\alpha',\beta').
\end{equation*}
We emphasise that top-$K$ Sinkhorn is a \emph{surrogate}, not an exact solver: the dual potentials $\mathbf f^{\star},\mathbf g^{\star}$ are obtained from the \emph{full} OT problem and then trimmed post-hoc, whereas solving partial OT directly requires modifying the Sinkhorn iteration with a dummy mass. Our shortcut does not provably solve the partial program, but it targets the same inlier-restricted scoring at the cost of a single $\mathcal{O}(N\log N)$ sort per pair, and \cref{tab:ablate-distance-method} shows it achieves comparable accuracy to the principled partial-OT solver of Chapel et al.~\cite{chapel2020partialOT} ($87.97$ vs.\ $87.97$ mIoU on our 135-pair validation set) at more than $100\times$ lower latency ($8$\,ms vs.\ $848$\,ms per pair). This keeps it inside the per-proposal candidate loop without breaking the runtime budget reported in \cref{tab:analysis-results}.

\myheading{Ground cost: cosine for global, squared-Euclidean for patch OT.}
The global distance $d_\mathrm{g}$ uses cosine distance to match the metric these vision--language encoders are trained under. The patch-level OT uses squared-Euclidean ground cost on $\ell_2$-normalised patch tokens, the native cost of the $p{=}2$ entropic Sinkhorn solver we use~\cite{cuturi2013sinkhorn,feydy2019sinkhorn}. This is consistent with cosine matching: after $\ell_2$ normalisation, the identity $\lVert u-v\rVert_2^2 = 2(1-u^\top v)$ shows that squared-Euclidean cost and cosine distance induce identical orderings on any pair of patches, so the choice does not change which patches the OT solver couples together. In pilot experiments using cosine on patches we observed no measurable accuracy difference, consistent with the orderings being identical.

\myheading{Per-matcher latency (details).}
The per-pair latencies reported in \cref{tab:ablate-distance-method} of the main paper are measured on an A100 GPU with $N{=}256$ BLIP-2 patches, averaged over the 135 \datasetname\ val pairs. The ordering follows directly from each matcher's computational structure. Top-$K$ Sinkhorn shares the same Sinkhorn forward pass as Full Sinkhorn, so the only added work is an $\mathcal{O}(N\log N)$ sort per side, empirically indistinguishable from the un-trimmed 8\,ms cost. Partial OT, in contrast, requires a network-simplex linear-programming solve (POT~\cite{chapel2020partialOT}) which does not parallelise on the GPU and is two orders of magnitude slower (848\,ms). Mutual-NN filtering adds an $\mathcal{O}(N^{2})$ pairwise argmax on top of Full Sinkhorn (10\,ms), so it is slightly more expensive while also degrading accuracy. This is why top-$K$ Sinkhorn is the only matcher in \cref{tab:ablate-distance-method} that attains partial-OT-level accuracy while remaining cheap enough for the per-proposal candidate loop.

\myheading{End-to-end latency breakdown.}
\cref{tab:latency-breakdown} decomposes the end-to-end inference time of \cref{tab:analysis-results} into its constituent stages, measured on a single A100 40\,GB GPU (same hardware as the rest of the paper) and averaged over the 135 \datasetname\ val pairs, with $B{\approx}19$ proposals and $N{\approx}1.6$ references per image; mask refinement is run only on the chosen box, not on all $B$ proposals. The dominant cost is foundation-model feature extraction (BLIP-2, $74\%$ of the budget), followed by SAM mask refinement ($21\%$). The proposed top-$K$ Sinkhorn matcher contributes only $1.5\%$ of total wall-clock time despite being the methodological centrepiece, confirming that the inlier-restricted matcher is essentially free at the system level. Conversely, this analysis identifies the encoder forward as the natural target for future deployment-side optimisation (e.g., distilled or smaller-resolution vision backbones), not the matcher.

\begin{table}[h]
\centering
\small
\caption{Per-stage latency breakdown of our pipeline on \datasetname\ val.}
\vspace{-4pt}
\begin{tabular}{l r r}
\toprule
\textbf{Stage} & \textbf{Mean (ms)} & \textbf{\% of total} \\
\midrule
Object detection & 34.8 & 3.4\% \\
Feature extraction & 749.2 & 73.7\% \\
\textbf{Matching (top-$K$ Sinkhorn)} & \textbf{14.8} & \textbf{1.5\%} \\
Mask refinement & 218.1 & 21.4\% \\
\midrule
\textbf{Total} & \textbf{1016.9} & \textbf{100\%} \\
\bottomrule
\end{tabular}
\label{tab:latency-breakdown}
\end{table}

\section{Evaluation Metrics}

We evaluate the performance of our method using Recall and Mean Intersection over Union (mIoU). These metrics assess both the quality of segmentation (PIS task) and the effectiveness of object proposals.

\subsection{Metrics for PIS Task}

\myheading{Recall}:  
For point-based methods such as PerSAM~\cite{persam}, PDM~\cite{pdm}, and P${^2}$SAM~\cite{p2sam}, Recall is computed by checking whether at least one predicted point prompt falls within the ground truth mask.  
For bounding box-based methods like ours, a prediction is considered correct if its Intersection over Union (IoU) with the ground truth mask exceeds 0.5.

\myheading{Mean Intersection over Union (mIoU)}:
To compute the mIoU for the PIS task, we first define the IoU for each category \( c \) as:

\begin{equation}
    \text{IoU}_c = \frac{|P_c \cap G_c|}{|P_c \cup G_c|}
\end{equation}

The overall mIoU is then calculated as the average IoU across all \( C \) categories:

\begin{equation}
    \text{mIoU} = \frac{1}{C} \sum_{c=1}^{C} \text{IoU}_c = \frac{1}{C} \sum_{c=1}^{C} \frac{|P_c \cap G_c|}{|P_c \cup G_c|}
\end{equation}

where:
\begin{itemize}
    \item \( P_c \) represents the set of predicted pixels for category \( c \),
    \item \( G_c \) represents the set of ground truth pixels for category \( c \),
    \item \( C \) is the total number of categories.
\end{itemize}

This metric evaluates segmentation quality by measuring the overlap between predicted and ground truth regions, with higher values indicating better performance.

\subsection{Object Proposal Evaluation (\cref{tab:analysis-results})}

\cref{tab:analysis-results} of the main paper reports proposal-stage recall \emph{before} any downstream matching. Because point-prompt and box-prompt methods produce different proposal types, each is scored under its native protocol:

\myheading{Recall (point-prompt baselines).}
For PerSAM~\cite{persam}, PDM~\cite{pdm}, and P$^2$SAM~\cite{p2sam}, a prediction is counted as a match if the predicted point prompt falls inside the ground-truth mask.

\myheading{Recall (our box-based proposals).}
For our method, a prediction is counted as a match if the proposal bounding box has IoU $>$ 0.5 with the ground-truth bounding box. Recall is then the percentage of ground-truth objects that have at least one matching proposal.

\myheading{Inference time.}
End-to-end runtime per target image, in milliseconds.

\myheading{Mean IoU (mIoU) of matched proposals.}
For all matched proposals, we compute the mean IoU as:

\begin{equation}
    \text{mIoU} = \frac{1}{N} \sum_{i=1}^{N} \frac{|P_i \cap G_i|}{|P_i \cup G_i|}
\end{equation}

where:
\begin{itemize}
    \item \( P_i \) is the predicted bounding box for object \( i \),
    \item \( G_i \) is the ground truth bounding box for object \( i \),
    \item \( N \) is the number of matched proposals.
\end{itemize}

This ensures that the evaluation considers both whether the reference object is reachable from the proposal stage (Recall) and the geometric quality of the matched proposals (mIoU).

\section{{\datasetname} Statistics}
\label{sec:dataset_suppl}

\begin{figure}[t]
    \centering
    \begin{subfigure}[t]{0.5\linewidth}
        \centering
        \includegraphics[width=\linewidth]{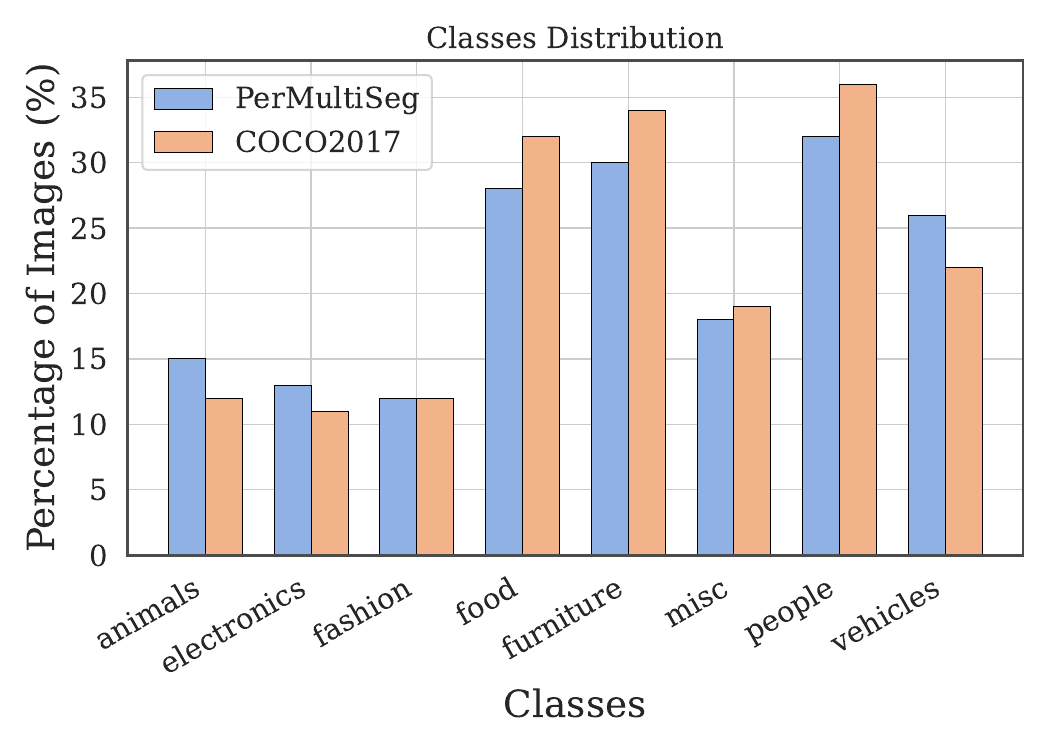}
        \caption{Classes distribution}
        \label{fig:class_dist_suppl}
    \end{subfigure}
    \hfill
    \begin{subfigure}[t]{0.49\linewidth}
        \centering
        \includegraphics[width=\linewidth]{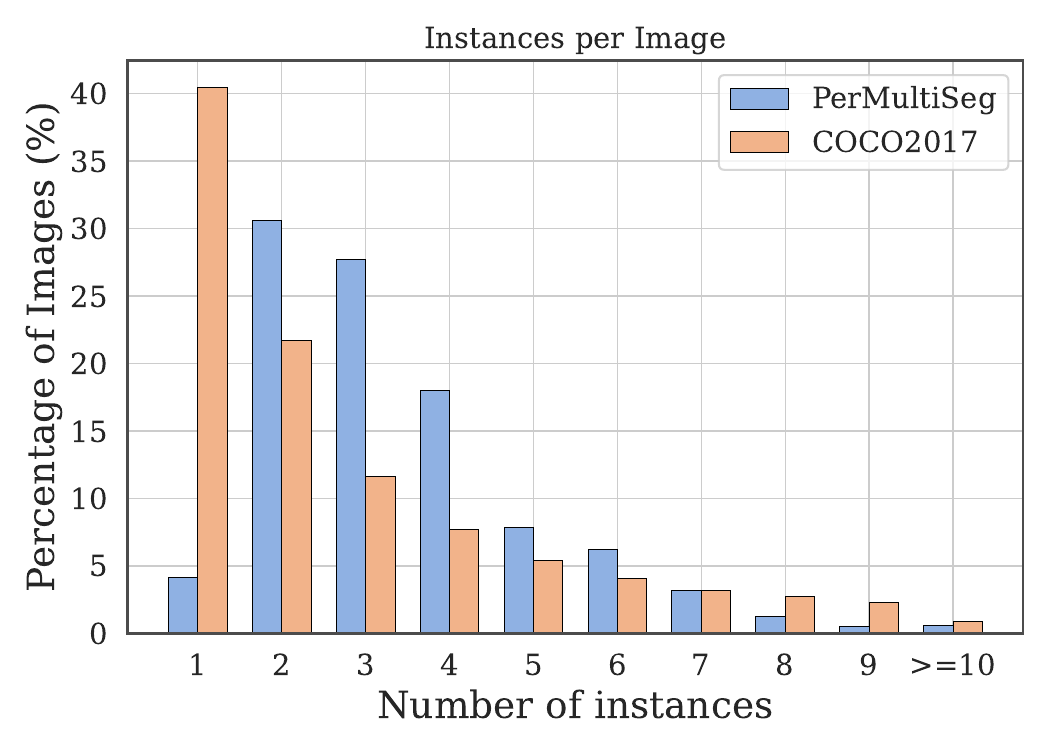}
        \caption{Instances per image}
        \label{fig:instance_per_image_suppl}
    \end{subfigure}
    \caption{Analysis of our \datasetname\ dataset vs.\ MS COCO.
    (a) Class distribution (\% of images per category).
    (b) Instances per image.
    \datasetname\ shows a more balanced class distribution and a higher frequency of multi-object scenes than COCO.}
    \label{fig:dataset-analysis-suppl}
\end{figure}

\myheading{Collection and annotation protocol.}
\datasetname\ consists of target images collected from both online sources and self-captured photographs, covering 45 categories with an average of 24--25 multi-object images per category. The images span diverse environments (indoor and outdoor), lighting conditions, camera viewpoints, and object types. As shown in \cref{fig:dataset-analysis-suppl}, \datasetname\ maintains a class distribution similar to MS~COCO while introducing significantly richer multi-object compositions: most target images contain two to four same-category instances (\cref{fig:instance_per_image_suppl}), increasing candidate-selection difficulty and placing higher demands on proposal recall and instance-level matching.

Overall, the dataset includes 1{,}079 target images and 2{,}147 reference images, with each target paired with roughly two references and 3{,}964 annotated instances in total (over three same-category objects per target on average). Each reference object is manually matched to its corresponding instance in the target, background-removed to eliminate noise, and annotated with a precise binary mask. This guarantees that every reference has a correctly labelled counterpart in the target, providing high-quality supervision for PIS evaluation.

\myheading{Instance-selection criterion.}
The difficulty of a PIS benchmark is set by how hard the distractors are, so we select them deliberately rather than accepting whatever same-category objects happen to be in a scene. For each target image we require the co-occurring instances to be the \emph{same product and the same colour} as the reference object, that is, visually near-identical units that differ only in physical identity, pose, wear, and placement. A distractor that is merely the same category (a different mug model, or the same mug in a different colour) is discriminable from a single global appearance vector, which is precisely the regime in which prior benchmarks saturate. Requiring same-product, same-colour distractors removes category- and colour-level shortcuts and forces the matcher to rely on fine-grained, spatially localised evidence such as labels, scuffs, contents, and surface detail. Each reference is annotated against exactly one instance in the target, so the ground truth remains unambiguous: the distractors share the reference's product and colour but never its identity. \cref{fig:hardcases_grid} shows representative cases.

\myheading{Category and scene bias.}
Two forms of bias are worth stating explicitly. \emph{Category bias:} the 45 categories are dominated by portable manufactured objects (mugs, bottles, bags, consumer electronics, footwear), because the same-product, same-colour criterion above requires multiple near-identical units to co-occur naturally, a condition mass-produced goods satisfy and organic or unique objects (animals, plants, artworks, faces) largely do not. \datasetname\ therefore measures instance discrimination in the manufactured-object regime and should not be read as covering deformable or organic instances. The per-category counts are nonetheless close to uniform (\cref{fig:class_dist_suppl}), avoiding the long-tailed frequency skew of MS~COCO. \emph{Scene bias:} images are weighted toward indoor, human-scale scenes (desks, shelves, tables, retail displays), a consequence of where multiple identical products naturally appear together; outdoor and long-range scenes are present but less frequent. Neither bias affects the comparisons in this paper, since all methods are evaluated on identical image and reference pairs, but both bound the external validity of the absolute numbers.

\begin{figure}[h]
\centering
\includegraphics[width=1\linewidth]{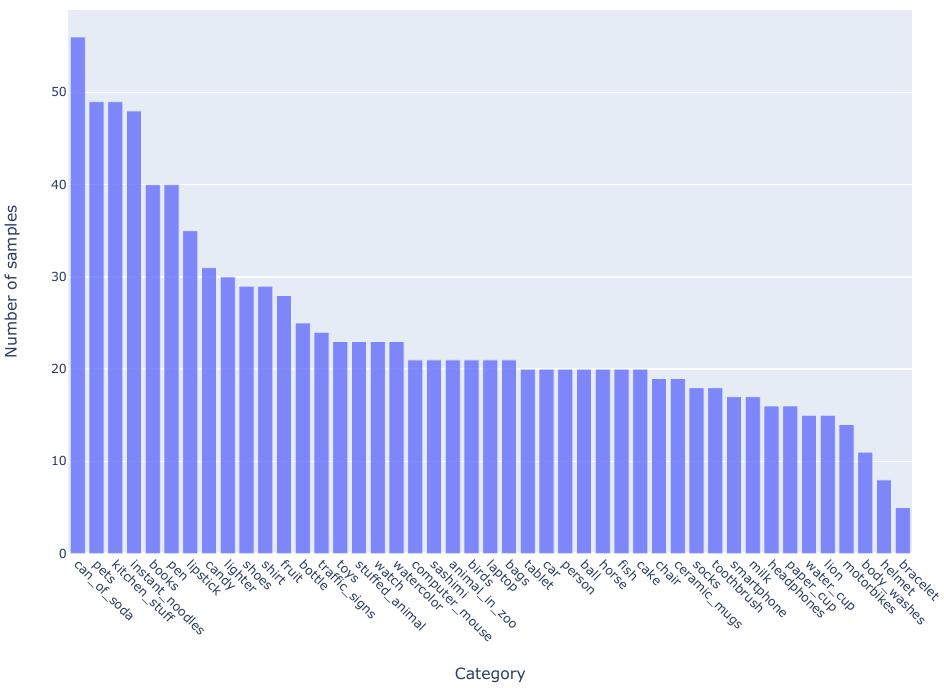}
\caption{The statistics of \datasetname, with the figure showing the category histogram.}
\label{fig:dataset-statistics}
\end{figure}

The category-level histogram is shown in \cref{fig:dataset-statistics}. Each category includes diverse object scenarios (varying environments, lighting conditions, camera views), and samples feature a range of resolutions and aspect ratios, making them more suitable for real-world applications than fixed-resolution benchmarks.

\section{Additional Ablations}

\myheading{Trim ratio $K$ (wider sweep).}
\cref{tab:ablate-alpha} sweeps $K\in\{0.10,\ldots,1.00\}$ on the default BLIP-2 configuration. Full Sinkhorn ($K{=}1$) is not the best; any $K\in[1/3,\,1/2]$ reaches the peak, and below $K{=}0.20$ the signal degrades as too few patches survive. We adopt $K{=}1/3$ in the main paper as the most aggressive trim at peak accuracy.

\begin{table}[h]
\centering
\small
\caption{Study on the trim ratio $K$ in top-$K$ Sinkhorn (BLIP-2, $\alpha_\mathrm{p}=0.75$).}
\vspace{-4pt}
\begin{tabular}{c c c c}
\toprule
\textbf{$K$} & \textbf{mIoU $\uparrow$} & \textbf{Recall $\uparrow$} & \textbf{bIoU $\uparrow$} \\
\midrule
0.10 & 85.60 & 87.38 & 80.40 \\
0.20 & 86.23 & 88.05 & 81.03 \\
0.25 & 86.67 & 88.50 & 81.47 \\
\textbf{1/3} & \textbf{87.97} & \textbf{89.83} & \textbf{82.69} \\
1/2 & 87.97 & 89.83 & 82.69 \\
3/4 & 86.67 & 88.50 & 81.47 \\
1 (full) & 85.04 & 86.83 & 79.93 \\
\bottomrule
\end{tabular}
\label{tab:ablate-alpha}
\end{table}

\begin{figure}[h]
\centering
\includegraphics[width=\linewidth]{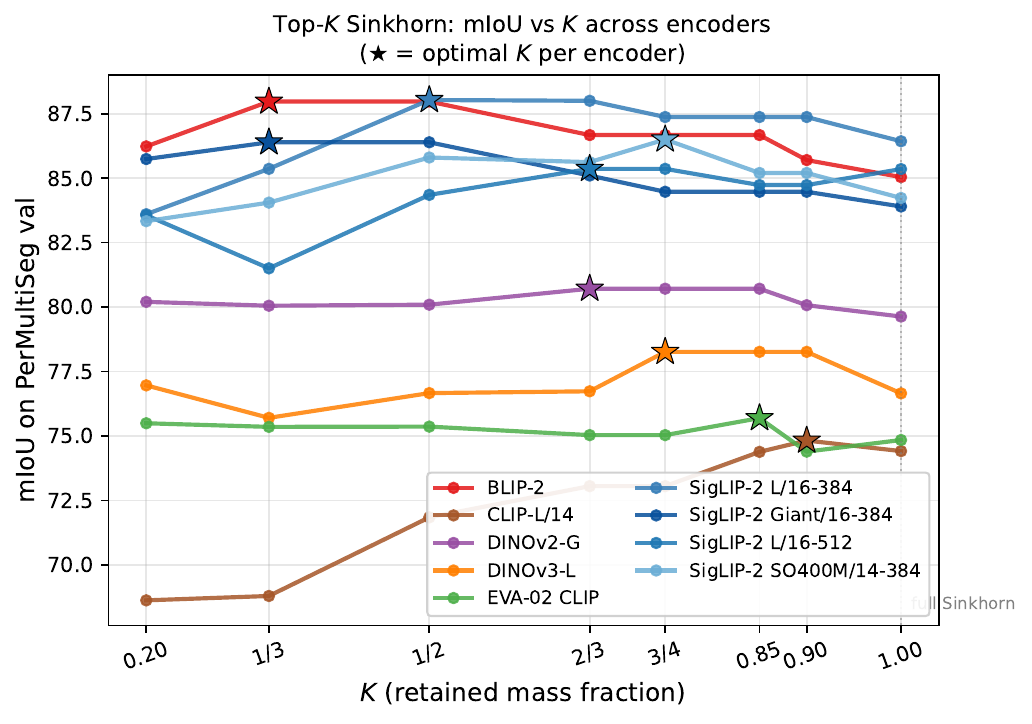}
\caption{Top-$K$ Sinkhorn $K$ sweep across encoders on \datasetname\ val.
$\bigstar$ marks the per-encoder optimal $K$. The dotted line at $K{=}1$
(full Sinkhorn) is suboptimal for every encoder. Optimal $K$ varies
across encoders ($K\!\in\![1/3,\,0.9]$) but trimming is universally beneficial.}
\label{fig:ksweep_encoders}
\end{figure}

\myheading{Per-encoder $K$ sweep.}
\cref{fig:ksweep_encoders} shows mIoU vs.\ $K$ for each encoder. \emph{Every encoder benefits from some degree of trimming} over full Sinkhorn ($K{=}1$): the gain ranges from $+0.4$ mIoU (CLIP-L) to $+2.9$ mIoU (BLIP-2), and trimming never hurts. The optimal trim ratio is mildly encoder-specific ($K^\star\!\in\![1/3,\,0.9]$), but the inlier-restricted scoring principle is universal. The strongest patch encoders (BLIP-2, the SigLIP-2 family) both benefit and peak at aggressive trims ($K^\star\!=\!1/3$ for BLIP-2 and SigLIP-2 Giant), consistent with the outlier-rejection view of top-$K$: discarding the highest-cost patches removes background and shared-category mass that otherwise dilutes the instance signal.

\myheading{Object proposal methods.}
\cref{tab:ablate-proposal} compares class-agnostic detectors as the proposal stage, on \datasetname\ and DAVIS2017. SAM's Automatic Mask Generator gives the highest recall on \datasetname\ but is far heavier and slower; YOLOv11 gives the best recall--accuracy--size trade-off across both datasets and is our default in the main paper.

\begin{table}[h]
\small
\centering
\setlength{\tabcolsep}{4pt}
\caption{Ablation on object proposal methods. Recall is proposal-level (box IoU $>$ 0.5) on the validation set.}
\vspace{-4pt}
\resizebox{\linewidth}{!}{
\begin{tabular}{l c c c c c}
  \toprule
  \textbf{Method} & \textbf{Params} & \multicolumn{2}{c}{\textbf{\datasetname}} & \multicolumn{2}{c}{\textbf{DAVIS2017}} \\
  \cmidrule(lr){3-4} \cmidrule(lr){5-6}
   &  & \textbf{mIoU $\uparrow$} & \textbf{Recall $\uparrow$} & \textbf{mIoU $\uparrow$} & \textbf{Recall $\uparrow$} \\
  \midrule
  SAM~\cite{sam} & 641M & \textbf{95.55} & \textbf{99.38} & 80.89 & 82.60 \\
  CODET~\cite{ma2023codet} & 45M & 94.09 & 98.51 & 80.37 & 83.79 \\
  YOLOW~\cite{yoloword} & 47M & 85.95 & 90.24 & 90.00 & 97.51 \\
  YOLOv11~\cite{khanam2024yolov11overviewkeyarchitectural} & 25M & 90.97 & 96.21 & \textbf{94.09} & \textbf{98.51} \\
  \bottomrule
\end{tabular}
}
\label{tab:ablate-proposal}
\end{table}

\myheading{Detection threshold sensitivity.}
\cref{tab:threshold} sweeps the YOLOv11 confidence threshold $\tau$ on the validation set (68 target images, 135 reference--target pairs). Lower $\tau$ yields more proposals, monotonically higher proposal recall, and monotonically higher final mIoU; the deployed $\tau{=}10^{-3}$ wins on every metric, reaching the same $87.97$ mIoU as the matcher ablation in \cref{tab:ablate-distance-method}. The extra proposals at low $\tau$ do not degrade matching: top-$K$ trims the high-cost outlier patches and per-pair matching latency is essentially flat across the sweep (17--19 ms) despite a $6\times$ increase in proposal count, so the recall--noise trade-off the threshold could expose is absorbed by the matcher rather than passed to the final mask.

\begin{table*}[t]
\small
\centering
\caption{Detection threshold sensitivity (\datasetname\ val, $\alpha_\mathrm{p}=0.75$, top-$K$ Sinkhorn at $K{=}1/3$). \emph{Proposal recall}: box IoU $>$ 0.5 with GT box. \emph{Recall}: output mask IoU $>$ 0.5.}
\vspace{-4pt}
\begin{tabular}{c c c c c c}
\toprule
\textbf{$\tau$} & \textbf{Avg props/img} & \textbf{Prop.\ Rec.\ $\uparrow$} & \textbf{Match (ms)} & \textbf{mIoU $\uparrow$} & \textbf{Recall $\uparrow$} \\
\midrule
$10^{-1}$ & 4.51  & 74.63 & 17.6 & 65.60 & 67.25 \\
$10^{-2}$ & 9.99  & 88.60 & 17.3 & 82.74 & 84.72 \\
\textbf{$10^{-3}$} & \textbf{27.69} & \textbf{91.42} & \textbf{18.9} & \textbf{87.97} & \textbf{89.83} \\
\bottomrule
\end{tabular}
\label{tab:threshold}
\end{table*}

\myheading{Self-term ablation (E5).}
The deployed top-$K$ Sinkhorn uses only the cross transport term $\mathrm{OT}_\epsilon(\V^r,\V^{(i)})$, omitting the two self-terms of the de-biased Sinkhorn divergence (\cref{eq:sinkhorn_div}). \cref{tab:selfterm} verifies this is lossless: adding self-terms yields identical mIoU at $3\times$ higher latency (three Sinkhorn solves instead of one).

\begin{table}[h]
\centering
\small
\caption{Self-term ablation: cross-only vs.\ de-biased $\mathcal{S}_\epsilon$.}
\vspace{-4pt}
\begin{tabular}{l c c c}
\toprule
\textbf{Config} & \textbf{$K$} & \textbf{mIoU} & \textbf{ms/pair} \\
\midrule
Cross-only & 1.00 & 85.04 & 8.3 \\
Cross-only & 1/3  & 87.97 & 7.9 \\
De-biased $\mathcal{S}_\epsilon$ & 1.00 & 85.04 & 22.9 \\
De-biased $\mathcal{S}_\epsilon$ & 1/3  & 87.97 & 22.7 \\
\bottomrule
\end{tabular}
\label{tab:selfterm}
\end{table}

\myheading{$\alpha_\mathrm{p}$ sweep (E3).}
\cref{tab:alpha_sweep} shows mIoU across blend weights $\alpha_\mathrm{p}$ for both full Sinkhorn and top-$K$. Top-$K$ helps at every $\alpha_\mathrm{p} \geq 0.75$ (up to $+2.93$ mIoU) and never hurts; the gain is not a hyperparameter accident. The advantage of top-$K$ over full Sinkhorn is largest where the patch distance carries the most weight: at the deployed default $\alpha_\mathrm{p}=0.75$ the gap is $+2.93$ mIoU ($87.97$ vs.\ $85.04$), narrowing to $+1.17$ at $\alpha_\mathrm{p}=0.85$ as the global CLS term begins to dominate the score. For completeness we also state the \emph{best-versus-best} comparison, which is the more conservative reading: full Sinkhorn peaks at $86.23$ ($\alpha_\mathrm{p}=0.50$) while top-$K$ peaks at $87.97$ ($\alpha_\mathrm{p}=0.75$), a gain of $+1.74$ mIoU. The larger $+2.93$ figure is measured at a shared $\alpha_\mathrm{p}$ where full Sinkhorn happens to dip below both of its neighbours, so $+1.74$ is the number we would defend as the matcher's own contribution. We adopt $\alpha_\mathrm{p}=0.75$ as the paper default, used consistently across all ablations.

\begin{table}[h]
\centering
\small
\caption{$\alpha_\mathrm{p}$ sweep: full Sinkhorn vs.\ top-$K$ Sinkhorn (mIoU).}
\vspace{-4pt}
\begin{tabular}{c c c c}
\toprule
$\alpha_\mathrm{p}$ & \textbf{Full Sinkhorn} & \textbf{Top-$K$} & \textbf{$\Delta$} \\
\midrule
0.00 & 83.86 & 83.86 & 0.00 \\
0.25 & 85.60 & 85.60 & 0.00 \\
0.50 & 86.23 & 86.23 & 0.00 \\
\textbf{0.75} & 85.04 & \textbf{87.97} & \textbf{+2.93} \\
0.85 & 85.83 & 87.00 & +1.17 \\
1.00 & 85.45 & 86.80 & +1.35 \\
\bottomrule
\end{tabular}
\label{tab:alpha_sweep}
\end{table}

\myheading{Entropic regularisation $\epsilon$.}
\cref{tab:entropic} shows that $\epsilon = 0.5$ yields the best balance; smaller values lead to unstable estimates while larger ones over-smooth the matching.

\begin{table}[h]
\centering
\small
\caption{Study on $\epsilon$ in entropic OT.}
\vspace{-4pt}
\begin{tabular}{c c c c}
\toprule
$\epsilon$ & \textbf{mIoU $\uparrow$} & \textbf{Recall $\uparrow$} & \textbf{bIoU $\uparrow$} \\
\midrule
0.1 & 86.76 & 89.05 & 81.35 \\
\textbf{0.5} & \textbf{87.39} & \textbf{89.72} & \textbf{81.93} \\
0.9 & 87.05 & 89.36 & 81.68 \\
1.0 & 86.76 & 89.05 & 81.35 \\
\bottomrule
\end{tabular}
\label{tab:entropic}
\end{table}

\myheading{Challenging conditions.}
\cref{tab:ablate_subset} stratifies by object count, occlusion, clutter, and illumination. Performance degrades under occlusion and many objects; clutter has modest effect.

\begin{table*}[t]
    \centering
    \small
    \caption{Performance under challenging conditions.}
    \vspace{-4pt}
    \begin{tabular}{lcccccccc}
        \toprule
        &  \textbf{[0--7]} & \textbf{[8{+}]} & \textbf{Occ} & \textbf{NoOcc} & \textbf{Clut} & \textbf{NoClut} &  \textbf{overexp} & \textbf{normal} \\
        \midrule
        \textbf{mIoU}   & 78.89 & 75.82 & 70.70 & 81.45 & 79.72 & 78.64 & 80.43 & 78.23 \\
        \textbf{Recall} & 80.89 & 79.34 & 72.58 & 75.71 & 81.85 & 80.67 & 74.59 & 80.34 \\
        \bottomrule
    \end{tabular}
    \label{tab:ablate_subset}
\end{table*}

\myheading{SAM family as image encoder.}
\cref{tab:sam_encoder} evaluates the vision encoders of SAM~\cite{sam},
SAM2~\cite{sam2}, and SAM3~\cite{sam3} as feature extractors for matching (not
as mask generators). Top-$K$ Sinkhorn improves over full Sinkhorn for all three
SAM variants, confirming that the inlier-restricted scoring principle is not
specific to vision-language encoders. SAM3's stronger semantic features yield
the highest absolute mIoU (80.25) among SAM-family encoders, but still fall
short of BLIP-2 (87.97) for PIS, since SAM-family encoders are trained for
mask prediction rather than instance discrimination.

\section{Additional Quantitative Results}
As few-shot segmentation (FSS) is closely related to personalized instance segmentation (PIS), we also evaluate FSS methods on the proposed PerMultiSeg dataset. As shown in \cref{tab:few-shots-method}, our method using a box prompt outperforms FSS methods by a significant margin. We also report results for our method using a center-point prompt, where the prompt is provided to SAM after candidate selection; this configuration still achieves better performance than prior methods.

\begin{table}[t]
\small
\centering
\setlength{\tabcolsep}{8pt}
\renewcommand{\arraystretch}{0.95}
\caption{Comparison on PerMultiSeg among few-shot semantic segmentation (FSS) methods and PIS methods.}
\vspace{-6pt}
\begin{tabular}{l l c c c}
  \toprule
  {\textbf{Method}} & \textbf{Task} & \textbf{mIoU $\uparrow$} & \textbf{Recall $\uparrow$} & \textbf{bIoU $\uparrow$} \\
  \midrule
  SegGPT~\cite{wang2023seggpt} & FSS & 29.34 & 16.86 & 21.23 \\
  Matcher~\cite{matcher} & FSS & 23.22 & 11.29 & 18.53 \\
  DCAMA~\cite{DCAMA} & FSS & 32.15 & 17.16 & 24.41 \\
  GF-SAM~\cite{GF-SAM} & FSS & 37.99 & 22.11 & 32.98 \\
  \midrule
  PerSAM~\cite{persam} & PIS & 21.65 & 25.29 & 14.82 \\
  PDM~\cite{pdm} & PIS & 23.65 & 43.91 & 12.82 \\
  P$^2$SAM~\cite{p2sam} & PIS & 31.58 & 24.57 & 27.90 \\
  \midrule
  Ours (point) & PIS & 55.41 & 55.20 & 48.66 \\
  Ours (box) & PIS & \textbf{80.07} & \textbf{82.29} & \textbf{75.15} \\
  \bottomrule
\end{tabular}
\label{tab:few-shots-method}
\end{table}

\section{Additional Qualitative Results}

\begin{figure*}[!t]
  \centering
  \includegraphics[width=1\linewidth]{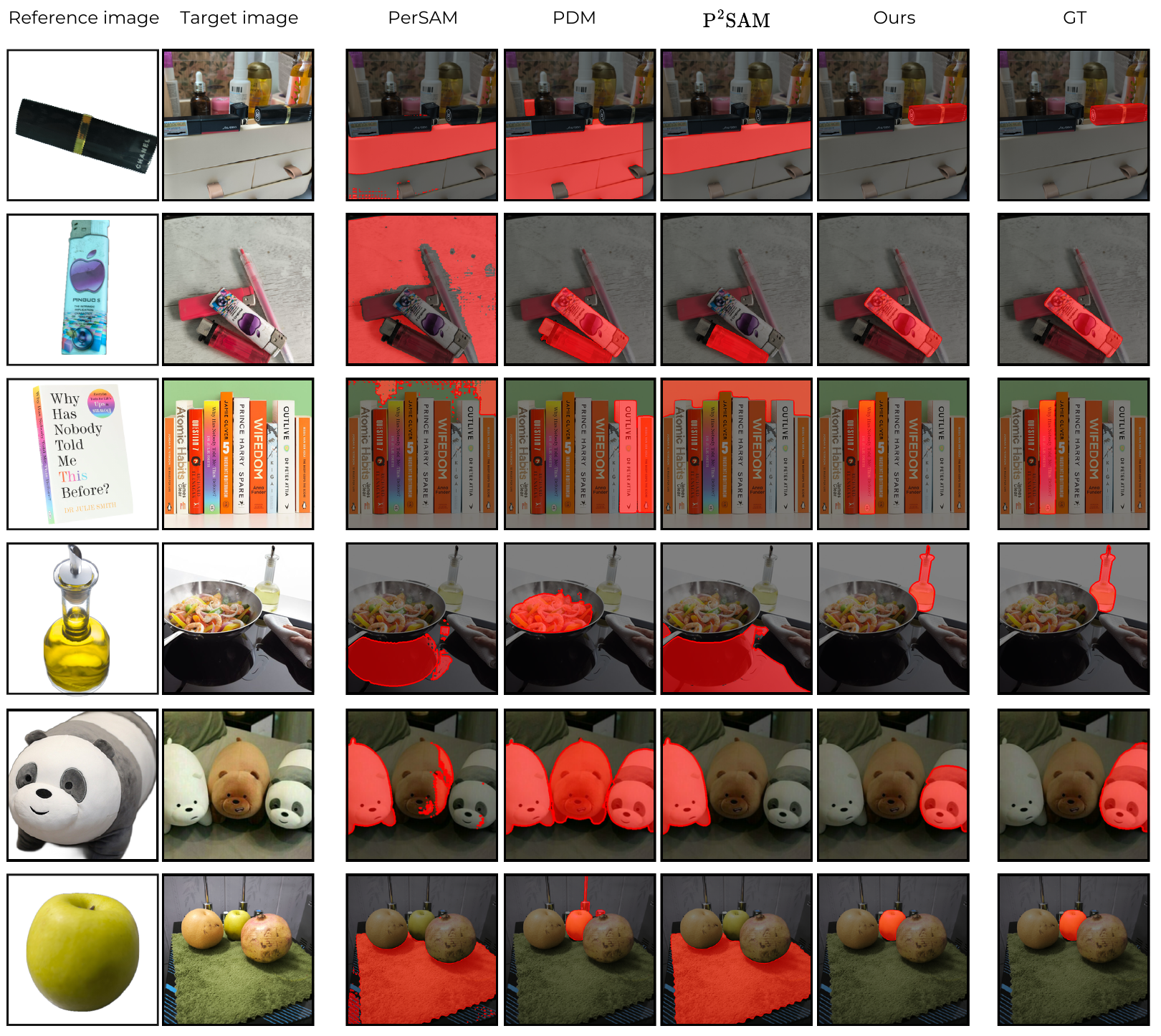}
     \caption{Additional qualitative examples on {\datasetname}.}
   \label{fig:qualitative_pms_extra}
\end{figure*}

We provide additional qualitative examples on \datasetname\ in \cref{fig:qualitative_pms_extra}. Our segmentation results outperform prior methods even in challenging cases such as the books (third row) and the fruit (last row).

\section{PIS vs.\ One-Shot Category Segmentation on LVIS}
\label{sec:lvis_discussion}

A natural question is why we do not evaluate on LVIS~\cite{lvis} or COCO~\cite{coco}, given that prior work such as PerSAM~\cite{persam} reports one-shot LVIS numbers. We address this directly rather than dismiss it. LVIS \emph{is} a valid one-shot benchmark, but it measures a different form of one-shot generalization: one-shot LVIS evaluates one-shot \emph{category} generalization (given one example of a category, segment \emph{all} its instances, scored by per-category AP), whereas PIS evaluates one-shot \emph{instance} generalization (given a reference image of one specific object, segment that same object among same-category look-alikes). These are complementary properties, not competing measurements of the same one: a method that segments all mugs scores well on LVIS but, on PIS, returns a distractor for every mug except the reference. Evaluating PIS on LVIS is possible in principle, but LVIS provides no reference-to-target instance correspondence (no ground truth for ``is this the same physical object as the reference?''), so it would require re-annotating correspondences and redefining the metric, yielding a \emph{new} benchmark rather than a comparison on LVIS as published. We instead build \datasetname\ with this correspondence infrastructure from the ground up, alongside the instance-level benchmarks PerSeg~\cite{persam}, PerMIS~\cite{pdm}, and DAVIS~\cite{davis2017}. That strong one-shot \emph{category} performance does not transfer to one-shot \emph{instance} discrimination is confirmed empirically: one-shot category-segmentation (FSS) methods reach only $23$--$38$ mIoU on \datasetname\ (\cref{tab:few-shots-method}), precisely because they cannot isolate the reference instance.

\section{Design Discussion}
\label{sec:design_discussion}

\myheading{Why training-free.}
None of the foundation models we compose (CLIP/BLIP-2 encoders, class-agnostic detectors, SAM) is trained on a personalized-segmentation objective, and building one is non-trivial: every new application domain (a specific user's belongings, a specific product catalogue, a specific industrial part) would require its own labelled reference--target instance correspondences and a tuned fine-tuning recipe, breaking the very generalisation that makes foundation models attractive. We therefore compose off-the-shelf detectors, encoders, and SAM into a personalized-segmentation pipeline with no per-domain training step. This keeps each component independently upgradable, so future stronger detectors or encoders translate directly into stronger PIS without re-collecting instance correspondences. Unified trained alternatives such as SAM3's Promptable Concept Segmentation~\cite{sam3} solve a related but different task (segment \emph{all} matching instances of a referenced concept, not the single specific instance corresponding to a reference image), as the SAM3 row of \cref{tab:permultiseg-results} shows empirically.

\myheading{Limitations and reference protocol.}
Our evaluation uses background-removed reference crops with a guaranteed correct match in the target, which is cleaner than many deployment scenarios where the user provides a phone screenshot, a loose crop, or a manual box, and the reference may not even appear in the target. Two failure modes follow directly: \emph{(i)} when no proposal covers the correct instance (detector recall, the bottleneck we already trade compute against in \cref{tab:ablate-proposal}), and \emph{(ii)} when the discriminative cue is occluded while a distractor displays a similar cue, as in the failure row of \cref{fig:qualitative_pms} of the main paper. Since the pipeline is modular and training-free, both are inherited from off-the-shelf components rather than baked-in: stronger class-agnostic detectors directly raise proposal recall, and the matcher accepts noisier reference crops to the extent the underlying encoder's CLS--patch fields remain instance-discriminative. Extending the benchmark to in-the-wild references (screenshots, loose crops, manual boxes) and adding a ``no-match'' rejection option are natural next steps.

\begin{table}[h]
\centering
\small
\caption{SAM family as image encoder for matching (\datasetname\ val).}
\begin{tabular}{l c c c c}
\toprule
\textbf{Encoder} & \textbf{CLS} & \textbf{Mean} & \textbf{Full} & \textbf{Top-$K$} \\
\midrule
SAM ViT-H~\cite{sam}    & 52.54 & 52.53 & 55.44 & \textbf{57.46} \\
SAM2 Hiera-L~\cite{sam2} & 61.15 & 59.19 & 57.94 & \textbf{59.08}  \\
SAM3~\cite{sam3} & 65.49 & 66.58 & 78.30 & \textbf{80.25}  \\
\bottomrule
\end{tabular}
\label{tab:sam_encoder}
\end{table}
\subsection{Baseline Configuration}
\label{sec:baseline_config}
For $\text{P}^2\text{SAM}$~\cite{p2sam} we report its official default configuration:
part-aware clustering enabled ($1 \le n_\mathrm{clusters} \le 5$), guided attention, and
the same SAM ViT-H checkpoint as all other methods. Running the baseline with a single
cluster disables its part-aware selection and understates it, so we use the stronger
setting throughout. Relative to the single-cluster setting, the official configuration
raises mIoU and bIoU but \emph{lowers} final-prediction Recall: it produces better masks
where it fires, yet clears the IoU $>$ 0.5 hit threshold less often.

\subsection{Where Does the Error Come From?}
\label{sec:error_decomposition}
We decompose end-to-end accuracy on the \datasetname\ test split ($1{,}971$ pairs) into
the two stages that can fail independently. A pair is \emph{recalled} if some proposal
box overlaps the ground-truth box with IoU $>0.5$, and \emph{correct} if the final mask
reaches IoU $>0.5$. Overall success approximately factorises as
$\mathrm{success} \approx \mathrm{ceiling} \times \mathrm{accuracy}\,|\,\mathrm{recalled}$
($91.58\% \times 85.98\% = 78.74\%$ vs.\ $79.55\%$ measured; the small residual reflects
mild positive correlation between proposal recall and matching accuracy across pairs).

\begin{table}[h]
\centering
\small
\caption{Error decomposition on \datasetname\ test ($1{,}971$ pairs).}
\vspace{-4pt}
\begin{tabular}{l c}
\toprule
\textbf{Quantity} & \textbf{Value} \\
\midrule
Proposal-recall ceiling & 91.58\% \\
Matching accuracy given recall & 85.98\% \\
Overall success & 79.55\% \\
\midrule
Unrecoverable (no proposal covers the object) & 8.42\% \\
Recalled but mismatched & 12.84\% \\
\midrule
mIoU when recalled & 82.68 \\
mIoU when the proposal is missed & 14.26 \\
\bottomrule
\end{tabular}
\label{tab:error_decomposition}
\end{table}

Two consequences are worth stating plainly. First, the pipeline has no recovery
mechanism: when the proposal stage misses the object, mIoU falls from $82.68$ to
$14.26$, so a detector miss is effectively terminal. Second, the residual error is
\emph{not} dominated by the early modules; $12.84\%$ of pairs are recalled but
mismatched against $8.42\%$ unrecoverable, so more loss currently sits in matching than
in proposal generation. This is why the matcher, rather than the detector, is the
component we ablate most carefully.

\subsection{Hard Cases: Clutter, Occlusion, and Near-Duplicates}
\label{sec:hardcases}
\begin{figure*}[t]
  \centering
  \includegraphics[width=\linewidth]{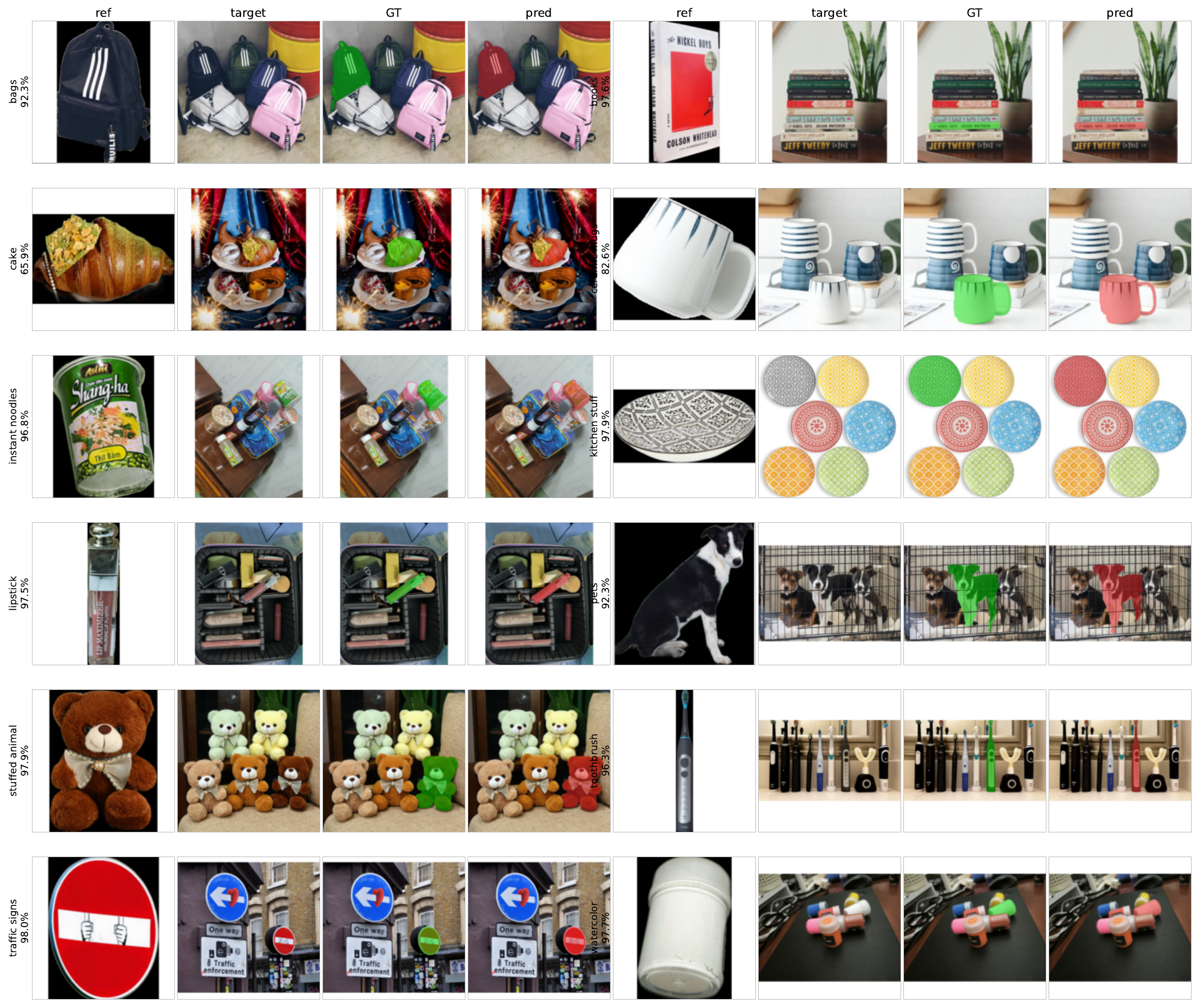}
  \caption{Hard cases on the \datasetname\ test split. Each block shows
  reference, target, ground truth (green), and our prediction (red), labelled with
  category and per-pair IoU. Rows cover dense same-category clutter, partial occlusion,
  and near-duplicate instances.}
  \label{fig:hardcases_grid}
\end{figure*}

\cref{fig:hardcases_grid} shows twelve test-split examples selected to span the
conditions that make \datasetname\ difficult: dense same-category clutter (bags, cakes,
mugs, instant noodles, kitchenware, lipsticks, watercolour sets), partial occlusion
(a book behind a plant stack, dogs behind cage bars), and near-duplicate
discrimination where instances differ only in a small region (traffic signs,
toothbrushes, stuffed animals). In each row the reference fixes a single instance among
visually similar candidates, and the prediction is shown against the ground truth.

\subsection{Statistical Reading of the Matcher Ablation}
\label{sec:matcher_stats}
Top-$K$ Sinkhorn and the partial-OT solver of Chapel et al.~\cite{chapel2020partialOT}
report identical values in \cref{tab:ablate-distance-method}
($87.97$ / $89.83$ / $82.69$). This is neither a coincidence nor a transcription error,
and the underlying agreement is stronger than the table alone conveys: the two matchers
select the same candidate box on \textbf{all 135} validation pairs, so their predicted
masks are identical and every downstream metric coincides exactly. Selection agreement,
rather than a coincidence of rounded averages, is what makes top-$K$ a usable stand-in
for the partial-OT solver it approximates at more than $100\times$ lower cost.
The same effect explains why neighbouring trim ratios share rows in
\cref{tab:ablate-alpha}: within the plateau $K\in[1/3,1/2]$ the trim does not change
which candidate wins.

We deliberately describe this as agreement on selections rather than statistical
equivalence. Because the prediction is decided by which candidate wins the argmin, mIoU
on this 135-pair split moves in discrete steps of roughly $0.7$ per flipped selection,
so the split is better suited to establishing that two matchers behave identically than
to resolving small differences between them. For the same reason we report the
conservative $+1.74$ best-versus-best gain over full Sinkhorn alongside the $+2.93$
measured at the shared default $\alpha_\mathrm{p}{=}0.75$ (\cref{tab:alpha_sweep}).

\subsection{Matched-Proposal Control: Giving Baselines Our Proposals}
\label{sec:matched_proposals}
A natural objection to \cref{tab:permultiseg-results} is that our pipeline consumes a
high-recall class-agnostic proposal set while the point-prompt baselines do not, so part
of the margin could reflect the input representation rather than the quality of
reference--candidate matching. We test this directly by giving
$\text{P}^2\text{SAM}$~\cite{p2sam} the \emph{same} proposals our system uses: instead of
selecting a point from its similarity map, it scores each of our candidate boxes with its
own reference embedding (mean-pooled over the in-mask reference patches) and the
highest-scoring box is refined by the same SAM checkpoint. Every other component is left
untouched.

\begin{table}[h]
\centering
\small
\setlength{\tabcolsep}{4pt}
\caption{Matched-proposal control on \datasetname. ``Matched'' replaces
$\text{P}^2\text{SAM}$'s point selection with a choice among our proposals, holding the
proposal set, SAM checkpoint, and its own features fixed.}
\vspace{-4pt}
\begin{tabular}{l l c c c}
\toprule
\textbf{Split} & \textbf{Variant} & \textbf{mIoU} & \textbf{Recall} & \textbf{bIoU} \\
\midrule
val  & $\text{P}^2\text{SAM}$ (official default) & 35.52 & 28.77 & 29.96 \\
val  & $+$ our proposals (mean pool) & 33.56 & 33.69 & 30.22 \\
val  & $+$ our proposals (max pool) & 32.04 & 32.69 & 29.28 \\
\midrule
test & $\text{P}^2\text{SAM}$ (official default) & 31.58 & 24.57 & 27.90 \\
test & $+$ our proposals (mean pool) & 22.11 & 22.15 & 20.04 \\
\bottomrule
\end{tabular}
\label{tab:matched_proposals}
\end{table}

Supplying the proposals does \emph{not} improve the baseline: it costs $1.96$ mIoU on
val and $9.47$ mIoU on test. High-recall proposals are therefore not a transferable
advantage on their own. Once the correct instance is present among many same-category
candidates, matching contributes beyond proposal recall: a single mean-pooled embedding
cannot separate look-alikes even when the correct box is available to select. The prompt
representation itself also matters (\cref{tab:few-shots-method} shows a center-point
variant of our system loses roughly half the gain), so the full margin reflects both
effects, not matching alone. The matching contribution is corroborated from the opposite
direction (\cref{tab:ablate-distance-method}), where replacing the patch-level distance
with mean-pooling costs more than 22 mIoU with the proposal set held fixed. We note that
this control favours the baseline on one axis, since it removes its point-selection stage
entirely and hands it an oracle-free but already-filtered candidate list.

\end{document}